\documentclass[10pt,twocolumn,letterpaper]{article}

\usepackage[pagenumbers]{cvpr} 

\usepackage[T1]{fontenc}
\usepackage[utf8]{inputenc}
\usepackage{soul}
\setuldepth{foobar}
\usepackage{tabularray}
\usepackage{tikz}
\usetikzlibrary{spy}
\usetikzlibrary{shapes.geometric}
\usepackage{pifont}
\usepackage{cuted}
\usepackage{stfloats}
\usepackage{placeins}

\definecolor{cvprblue}{rgb}{0.21,0.49,0.74}
\usepackage[pagebackref,breaklinks,colorlinks,allcolors=cvprblue]{hyperref}

\def\confName{CVPR}
\def\confYear{2027}

\def\ourB{SAPER\includegraphics[width=0.7em]{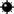}}
\def\our{SAPER}

\newcommand{\xmark}{\ding{55}}%
\usepackage{mathtools}

\title{Pruning as a Lens: Understanding the Functional Roles of Attention Heads in Vision Transformers}

\author{
  Kamil Książek$^{1}$, Michał Jan Włodarczyk$^{*,1}$, Piotr Suszyński$^{*,1}$, Jacek Tabor$^{1,2}$, Przemysław Biecek$^{1,3}$ \\[4pt]
  \small $^1$Centre for Credible Artificial Intelligence, Warsaw University of Technology, Poland \\
  \small $^2$Faculty of Mathematics and Computer Science, Jagiellonian University, Krakow, Poland \\
  \small $^3$University of Warsaw, Poland \\[3pt]
  \ttfamily\footnotesize \{kamil.ksiazek, piotr.suszynski.dokt, michal.wlodarczyk, przemyslaw.biecek\}@pw.edu.pl \\
  \ttfamily\footnotesize jacek.tabor@uj.edu.pl
}

\begin{document}
\maketitle
\begin{figure*}[!t]
    \centering
    \includegraphics{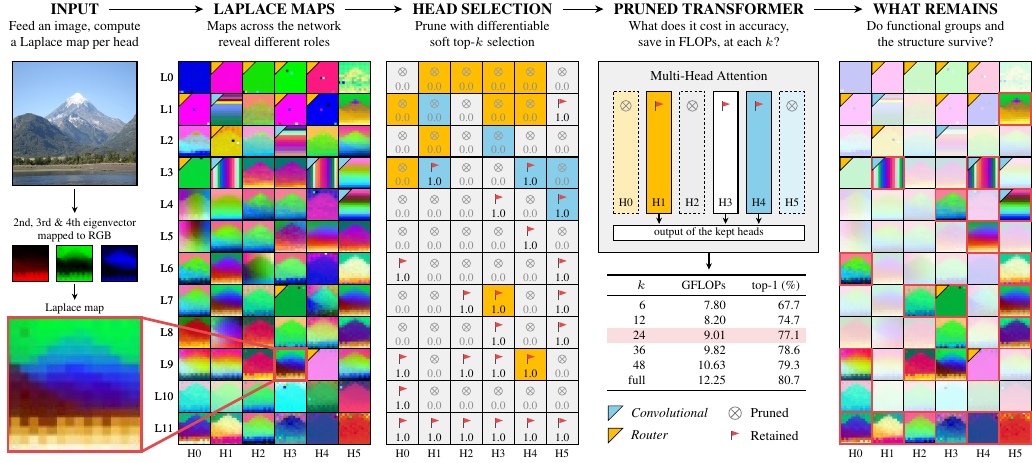}
    \caption{\textbf{Overview of \ourB{}}. For an input image, every attention head of DINOv2~\cite{oquab2024dinov} is visualized as a Laplace map, which reveals the head's role. On the left side, a single head and its 2nd to 4th smallest Laplacian eigenvectors mapped to RGB are presented. A differentiable soft top-$k$ operator retains $k$ heads, and the rest are pruned. We then measure accuracy and FLOPs, and study which heads are retained. The budget $k$ is varied throughout the paper. Also, the grids show ViT-S with a $k=24$-element subset of its 72 heads.}
    \label{fig:teaser}\label{fig:semantic_clustering}
\end{figure*}

\begin{abstract}
Vision foundation models, such as DINOv2, learn highly expressive representations but rely on massive, opaque architectures that demand substantial computational and memory resources. To provide an interpretability-guided and efficient solution to this issue, we first propose Laplace maps, a novel spectral analysis and visualization technique for individual attention heads based on the Laplacian eigenvectors of attention activations. This allows us to categorize the functional specializations of individual heads, uncovering local convolutional processing, topological routers, and complex semantic parsing. Building upon recent observations regarding contiguous representational phases in vision transformers, we extend this discovery to the head level, assess their similarity, and identify functional redundancies. Leveraging these insights, we introduce Soft Attention PrunER (SAPER), an end-to-end differentiable pruning framework powered by the LapSum Soft top-k operator. By clearing redundant heads, SAPER significantly reduces FLOPs while preserving core model performance, allowing us to quantitatively assess the architectural importance of router heads under constrained capacity.

\end{abstract}

\section{Introduction}

Vision transformers (ViTs)~\cite{dosovitskiy2021image}, and in particular self-supervised foundation models such as DINOv2~\cite{oquab2024dinov}, have become the backbone of a wide range of downstream applications, from image classification to segmentation and dense prediction. Their success, however, comes at a price: these models are heavily over-parameterized, and the multi-head self-attention mechanism~\cite{vaswani2017attention} contributes to both a computational bottleneck and an increased opaqueness. Each layer hosts a set of attention heads that jointly transform the representation, yet it remains unclear what specific role each head plays, whether functional redundancies exist, and how many heads are strictly necessary to preserve full-model performance. As foundation models migrate from data centers to edge devices, this question is a highly practical one, every retained head translates into latency, memory, and energy consumption, making structured, head-level compression a natural instrument of the Green AI agenda~\cite{schwartz2020greenai,strubell2019energy}.

Importantly, the redundancy we target is not a defect of the
architecture but an emergent consequence of how foundation models are
trained. Self-supervised pre-training on millions of
images~\cite{oquab2024dinov} rewards the network for maintaining many parallel computational pathways: features useful for one downstream
task may be irrelevant for another, and it cannot be determined a priori which specific features will be needed. The result is a general-purpose
representation in which, for any {specific} task, a substantial
fraction of attention heads is functionally duplicated or inactive.
From this perspective, pruning is a {specialization} of a general model rather than a repair of a flawed one. Moreover, determining the optimal network substructure is easier after finishing the training, as the lottery ticket hypothesis suggests~\cite{frankle2018lotteryticket}.

While model redundancy is a problem, blind reduction of the model, taking into account only the model's performance and size, is not always the best solution. Pruning should instead be grounded in an understanding of what the removed components actually compute. This view aligns with a broader shift in the community: the call for a rigorous science of model interpretability~\cite{doshivelez2017rigorous}, the agenda of mechanistic interpretability, which seeks to reverse-engineer the roles of individual model components such as attention heads~\cite{sharkey2025open}, and, most recently, the program of Model Science~\cite{biecek2026modelscience}, which postulates studying trained models as objects of systematic inquiry.

Our work instantiates this program for the attention mechanism of vision foundation models: we first explore the functional structure of individual attention heads, and only then refine the model by removing the heads, which this analysis confirms to be redundant. The outline of \our{} is presented in Figure~\ref{fig:teaser}. We approach the explainability of attention heads from a graph-theoretic perspective. An
attention map can be interpreted as a weighted graph over tokens, and
recent work has analyzed such graphs as discrete-time Markov
chains~\cite{erel2026attention} or leveraged the spectrum of their
Laplacians as a diagnostic signal~\cite{binkowski2025hallucination,
wang2024zerotprune}. Building on this view, in our paper we address the following questions:

\begin{enumerate}
    \item \textbf{What is the functional role of individual attention heads?} We introduce Laplace maps, a spectral-based technique that characterizes head specialization. This allows us to explicitly identify \emph{convolutional} heads, \emph{router} heads, and heads dedicated to extracting complex semantic patterns.
    \item \textbf{How do attention heads interact structurally?} Using Laplacian-based spectral clustering, we generalize the recent discovery of contiguous representational phases~\cite{jacobs2026raptor} to the structural level of individual attention heads.
    \item \textbf{Which heads drive model performance?} We propose \ourB{}\footnote{The name \ourB{} is inspired by the game of Minesweeper: a head's true importance is hidden, but its spectral signature provides indirect evidence. This allows us to safely clear redundant heads while leaving untouched the critical \emph{mines}, i.e. heads whose removal would collapse the representation.}, an end-to-end differentiable head pruning framework powered by the LapSum soft top-$k$ selection operator. We perform an in-depth analysis of DINOv2 and evaluate the structural significance of \emph{router} heads.
\end{enumerate}

\section{Related works}
\subsection{Interpretability of attention heads in ViTs}
One of the prominent directions in vision transformer research is decoding the internal mechanics of attention. Self-supervised ViTs inherently develop object-centric attention~\cite{caron2021dino}, with heads specializing into distinct functional roles that emerge during training rather than through architectural priors~\cite{raghu2021vision, voita2019analyzing}. A large body of attribution methods seeks to explain ViT predictions by propagating relevance across layers, utilizing Attention Rollout~\cite{Abnar2020LRPbased}, Deep Taylor Decomposition~\cite{chefer2021transformer}, or AttnLRP~\cite{achtibat2024attnlrp}. However, these approaches are predominantly label-dependent. In contrast, \our{} characterizes the intrinsic, label-independent topological structure of a head's attention graph, yielding a universal behavioral signature for each head.

\subsection{Graph-theoretic and spectral analysis}
Recent works increasingly formalize the attention matrix as a graph to analyze its properties spectrally. Zero-TPrune~\cite{wang2024zerotprune} and TokenRank~\cite{erel2026attention} leverage stationary distributions and Markov chain interpretations for training-free token pruning or semantic grouping. Spectral features of attention maps have also served as diagnostic signals, such as for hallucination detection~\cite{binkowski2025hallucination}. \our{} adapts this graph-theoretic view specifically for head-level interpretability. We compute the Laplacian spectrum of each head, mapping the Fiedler vector~\cite{shi2000normalized, Liu2011laplacian} and subsequent higher-order eigenvectors to RGB spatial maps (Laplace maps) to explicitly visualize partitioning behavior. While DSMI~\cite{Shivanvitha2024spectral} similarly extracts higher-order eigenvectors, it relies on post-hoc, gradient-weighted combinations strictly to visualize visual-semantic alignment in frozen vision-language models. \our{} instead uses these spectral signatures to structurally categorize heads (e.g. \emph{router} or \emph{convolutional}) and explains their significance within the network.

\subsection{Pruning as structural validation}
To empirically validate these spectral discoveries, we leverage network pruning not merely as a compression tool, but as an identification of structural redundancy. Traditional ViT compression targets tokens, channels, or heads to maximize FLOP reduction, utilizing global $L_1$ sparsity penalties (ViT-Slim~\cite{chavan2022vision}), memory-intensive stochastic evolutionary search as in SaViT~\cite{chuanyang2022savit}), or target-class gradient masking (X-Pruner~\cite{lu2023xpruner}). Other approaches focus on sequence merging, such as ToSA~\cite{huang2025tosa}, joint reduction of attention heads and tokens or MLPs, e.g. HEART-ViT~\cite{uddin2025heartvit}, UP-ViT~\cite{Yu2023unifiedpruning}, or Gumbel-Softmax top-$k$ relaxations for NLP heads~\cite{li2021differentiable}. SAPER, as an entire pipeline, not only selects the most important attention heads, reducing FLOPs, but also analyzes the structural behavior of the model, as well as identifies and uncovers the role of \emph{router} heads.

\begin{figure*}[!ht]
    \centering
    \begin{subfigure}{0pt}\phantomsubcaption\label{fig:head_similarity}\end{subfigure}%
    \begin{subfigure}{0pt}\phantomsubcaption\label{fig:head_clusters}\end{subfigure}%
    \begin{subfigure}{0pt}\phantomsubcaption\label{fig:PCA}\end{subfigure}%
    \includegraphics{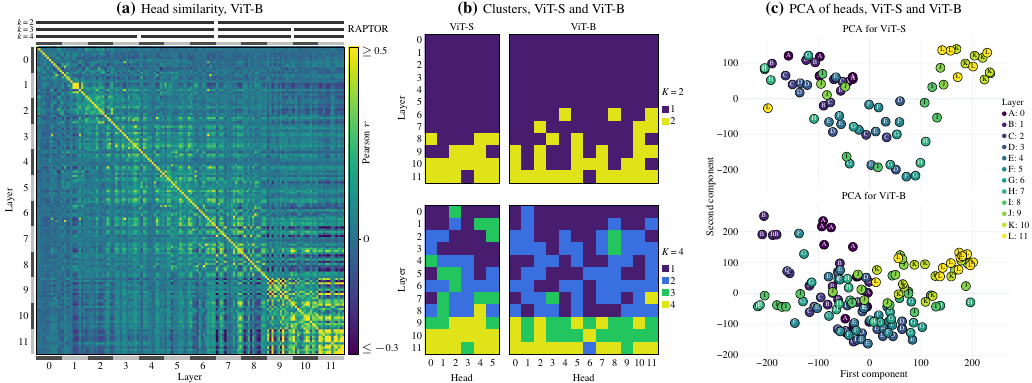}
    \caption{\textbf{Structure of the attention heads of DINOv2 on ImageNet-1K.} \textbf{(a)}~Pearson correlation between the heads of ViT-B, ordered by layer; the colour scale is clipped at $-0.3$ and $0.5$. The bars above the matrix show the contiguous blocks that RAPTOR~\cite{jacobs2026raptor} reports for $k = 2, 3, 4$ (for ViT-B with registers). \textbf{(b)}~Semantic clustering of the ViT-S and ViT-B heads into $K = 2$ and $K = 4$ clusters. \textbf{(c)}~First two principal components of the head activations of ViT-S and ViT-B, coloured by layer.}
    \label{fig:placeholder}
\end{figure*}

\section{Methods}
\subsection{Semantic clustering}
Through semantic clustering, we identify functionally similar attention heads across varying network depths. Unlike RAPTOR~\cite{jacobs2026raptor}, which enforces contiguous block partitioning, our head-level assignment relies strictly on learned representations, uncovering a more subtle, non-sequential model topology.

For each image $i$ from the $N$-element dataset, and head $h \in \lbrace 1, ..., H \rbrace$ in an $L$-layer network ($H=72$ for DINOv2 ViT-S and $H=144$ for ViT-B, $L=12$ in both cases), we extract the pre-softmax attention logits $\mathbf{A}_i^h \in \mathbb{R}^{D \times D}$, where $D$ is the sequence length. To capture global context aggregation, we compute the softmax distribution over the entire sequence from the \emph{CLS} token (index $0$) and discard its self-attention, yielding the spatial attention vector $\mathbf{p}_i^h \in \mathbb{R}^{D-1}$, i.e. $\mathbf{p}_i^h[j] = \frac{\exp(\mathbf{A}_i^h[0, j])}{\sum_{k=0}^{D-1} \exp(\mathbf{A}_i^h[0, k])}$, where $j \in \{1, \dots, D-1\}$.

Extracting the peak spatial attention $\max_j \mathbf{p}_i^h[j]$ across all $N$ images yields an $N$-dimensional feature vector per head. The pairwise Pearson correlations between them form an $H \times H$ similarity matrix $S$, which we partition into $K$ clusters through spectral clustering. Mathematically, it is equivalent to solving the normalized cut problem~\cite{Liu2011laplacian} or clustering Markov chain states~\cite{Meila2000randomwalks}. Finally, to visualize inter-head relationships in 2D, we concatenate the maximum, mean, and standard deviation of $\mathbf{p}_i^h$ across all $N$ images into a feature matrix $\mathbf{P} \in \mathbb{R}^{H \times 3N}$ and apply PCA (or joint PCA on the combined head sets when comparing across distinct ViT architectures).

\subsection{Laplace maps}
An attention map for the head $h$ and the image $i$, $\mathbf{A}_i^h \in \mathbb{R}^{D \times D}$, where $D$ denotes the total number of tokens, including the CLS token, can be interpreted as a weighted graph $\mathbf{G} = (V, E)$. Vertices $V$ represent individual tokens, and edges $E$ (and their corresponding weights) encode pairwise attention strengths. Graph-theoretic tools enable rich analysis of these interaction patterns. For instance, recent work considered attention matrices as discrete-time Markov chains to study convergence to stationary states~\cite{erel2026attention}. Here, we propose a graph spectral visualization method for attention maps to better understand the specific roles of individual attention heads across dataset samples.

We first apply softmax normalization and drop the CLS token row and column from $\mathbf{A}_i^h$, yielding the spatial patch adjacency matrix $\hat{\mathbf{A}}_i^h \in \mathbb{R}^{(D-1) \times (D-1)}$. To construct an undirected graph representation, we symmetrize this matrix through $\hat{\mathbf{A}}_i^{h, \text{sym}} = \frac{1}{2} \left( \hat{\mathbf{A}}_i^h + (\hat{\mathbf{A}}_i^h)^\top \right)$. Next, we compute the degree matrix $\mathbf{D}_i^h$, representing the average of both incoming and outgoing attention weights for consecutive nodes. The unnormalized graph Laplacian is then given by $\mathbf{L}_i^{h, \text{sym}} = \mathbf{D}_i^h - \hat{\mathbf{A}}_i^{h, \text{sym}}$, which is symmetric positive semi-definite matrix. We extract the eigenvectors corresponding to the second, third and fourth smallest eigenvalues and normalize the resulting coordinates. Crucially, the eigenvector associated with the second smallest eigenvalue of the Laplacian, often called the Fiedler vector, provides continuous relaxation to the balanced minimum cut problem~\cite{Liu2011laplacian, von2007spectral}. Finally, we construct an RGB visual map by assigning these three eigenvectors to the red, green, and blue channels, respectively. As an ablation study, we also evaluate visualizations derived from higher-order eigenvectors and present them in Appendix A.

In principle, selecting eigenvectors reduces to the minimization of the Laplacian quadratic form $\mathbf{x}^{\top}\mathbf{L}\mathbf{x} = \frac{1}{2} \sum\nolimits_{m,n=1}^{D} \mathbf{A}_{m,n} (x_m - x_n)^2$, where image and head indices are omitted for clarity. Minimizing this objective ensures that token pairs connected by large attention weights are assigned similar scalar values $x_m \approx x_n$. The smallest eigenvalue yields a trivial constant eigenvector. Therefore, we select the next three consecutive eigenvectors, capturing graph global topology, low-frequency structural components, and optimal spatial token clustering~\cite{ortega2018graphs,shuman2013graphs}.

\subsection{Differentiable attention head selection}
LapSum~\cite{struski2025lapsum} formulates a differentiable relaxation of the discrete (hard) top-k selection problem, referred to as the soft top-k operation. In hard top-k, given a score vector $r = (r_0, r_1, ..., r_{H-1}) \in \mathbb{R}^H$, one creates a $H$--dimensional binary mask $\hat{p} \in \lbrace 0, 1 \rbrace ^H$. In this mask, ones correspond to the highest $k$ scores in $r$, and zeros to the remaining $H-k$ elements. Equivalently, this establishes a strict threshold between the $k$--th and $(k+1)$--th sorted elements of $r$. In the hard top-k, $k$ is restricted to integer values, while in the soft top-k we can select an arbitrary continuous budget $k \in (0, H)$. Therefore, it produces a soft mask whose elements sum to a total mass $k$. LapSum computes the soft selection probabilities $p = (p_0, p_1, ..., p_{H-1})$ such that the $i$--th component ($i \in \lbrace 0, 1, ..., H-1 \rbrace$) is given by:

\begin{equation}
	p_i = LapSum(r_i, \alpha) = \mathrm{Lap}\left( \dfrac{b - r_i}{\alpha} \right),
\end{equation}

\noindent where $p_i \in [0, 1]$ represents the soft selection weight for the $i$--th score ($r_i$) under the constraint $\sum_{i=0}^{H-1} p_i = k$. The parameter $\alpha < 0$ acts as a negative temperature, and $\mathrm{Lap}(\cdot)$ represents the cumulative distribution function (CDF) of the Laplace distribution:

\begin{equation} 
	\mathrm{Lap}(x)  =\left\{
	\begin{array}{@{}ll@{}}
		\frac{1}{2} \exp(x) & x \leq 0\\
		1 - \frac{1}{2} \exp(-x) & x > 0.
	\end{array}\right.
\end{equation}

As $\alpha \rightarrow 0^{-}$, the soft mask sharply approaches the Heaviside step function, recovering the exact hard top-k ranking of the scores. The scalar barrier $b$ is determined through an inverse-CDF query on a smoothed cumulative curve defined by the sorted scores. LapSum is leveraged in our method to select the top-k attention heads in an end-to-end differentiable manner. Its time complexity is equal to $\mathcal{O}(H\cdot \log(H))$.

\section{Experiments}
\subsection{Procedure} 
\begin{figure*}[t]
    \centering
    \includegraphics{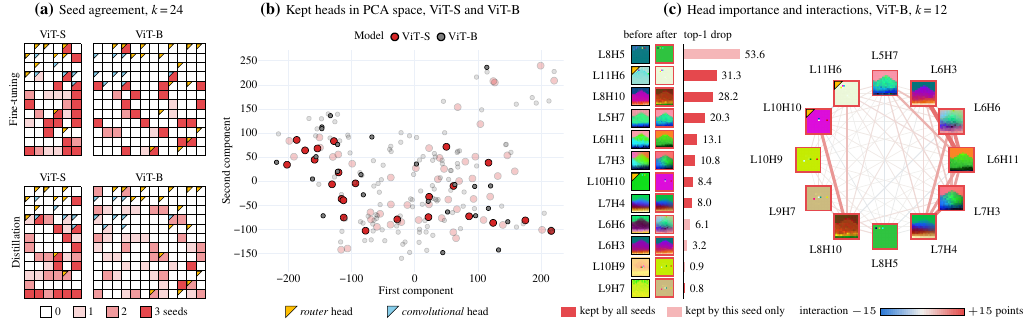}
    \caption{\textbf{Pruned models on ImageNet-1K.} \textbf{(a)} Head selection counts across runs ($k=24$) after fine-tuning (top) and distillation (bottom). All runs agree on 18 and 20 heads (top) and on 14 and 2 (bottom) of ViT-S and ViT-B, respectively. \textbf{(b)} Joint PCA of ViT-S and ViT-B head activations, where semi-transparent markers denote heads pruned in a single \our{} run. \textbf{(c)} Ranked individual head accuracy impact for a fine-tuned ViT-B ($k=12$), shown alongside pre- and post-pruning Laplace maps and pairwise head interactions (line thickness indicates pair importance).}
\label{fig:pruned_models}\label{fig:pca_selected_heads}\label{fig:commonPCA}\label{fig:selected_heads_laplace}
\end{figure*}

We assessed the relationship between various model attention heads through semantic clustering using $N=50,000$ images sampled from the ImageNet-1K validation set. We present a closer look at the Laplace maps for the selected images from the same dataset.

Regarding attention head pruning experiments, we selected RAPTOR~\cite{jacobs2026raptor} as our main baseline because we are extending their work. RAPTOR identified a block structure in ViTs and replaced the original pre-trained blocks with a smaller number of recurrent block partitions. However, in our case, blocks are not necessarily contiguous, but individual heads are assigned to one of the clusters. In the case of RAPTOR, the minimum number of heads is equal to $6B$ for ViT-S and $12B$ for ViT-B, where $B$ is the number of selected blocks. The authors consider only at least 2 blocks. Our method enables the use of any number of attention heads $k \in \lbrace 1, ..., H \rbrace$, where $H$ is the total number of heads ($H=72$ in ViT-S and $H=144$ in ViT-B). We also included two simple baselines: direct mask and magnitude pruning. For direct mask, we assign a trainable parameter to each head, apply a sigmoid function and multiply the head output by the obtained score. During training, we add two components to the cross-entropy loss function: one pushes consecutive mask values $m_i$ to binary values $\frac{1}{n} \sum\nolimits_{i=1}^{n} m_i \cdot(1 - m_i)$~\cite{murray2010optimization}, and the second minimizes the $L_2$-norm of the difference between the sum of the proposed masks and $k$. Similarly to \our{}, we gradually decrease the number of remaining heads. For magnitude pruning, we just calculate the Frobenius norm of the attention head activations averaged across all data samples.

To ensure a reliable comparison with other methods, we performed experiments on DINOv2 ViT-S and ViT-B~\cite{oquab2024dinov}, focusing on ImageNet-1K~\cite{deng2009imagenet} and CIFAR-100~\cite{Krizhevsky09cifar} (full results in Appendix F). We evaluated \our{} in two modes: pure fine-tuning or knowledge distillation from the initial DINOv2 model. The selected model is initially trained for $N_s$ epochs with simultaneous soft top-k head selection, then trained for $N_h$ consecutive epochs only with the $k$ most important attention heads according to the previous option (hard top-k stage). With pure fine-tuning, the model is trained with a linear classification probe and utilizing the cross-entropy loss, while in knowledge distillation, the loss is the cosine similarity between the outputs of the selected layers of the teacher and corresponding student model activations after a linear projection. The details of hyperparameters are described in Appendix G.

\subsection{Results}

\paragraph{Semantic clustering}
Our semantic clustering results align with the observations of RAPTOR, revealing the block structure across both ViT variants, see Figure~\ref{fig:placeholder} and Appendix B.1. In ViT-S, we identified a clear functional distinction between heads in layers L0-L7 and those in L8-L12. Notably, in the two-cluster mode, individual heads within the last four layers, surrounded by heads from cluster 2, retain high similarity to early-layer representations, mainly assigned to cluster 1. A similar observation can be made for ViT-B, characterized by a \emph{continuous transition} between L8 and L9. Increasing the cluster granularity further narrows the distinct late-stage head representations to the final two to three layers of ViT-S. While specific head assignments vary slightly between CIFAR-100 and ImageNet-1K, the overall structural dynamics remain comparable across datasets. Crucially, our findings generalize the block-level discoveries of RAPTOR by shifting the analysis from entire layers to individual attention heads, demonstrating that functionally analogous heads frequently emerge across different network depths.

\paragraph{Principal component analysis} In ViT-S, we observe a smooth trajectory across consecutive layers, as illustrated in Figure~\ref{fig:PCA}. However, individual heads in the final layers exhibit greater similarity to early-layer heads than to their immediate neighbors. For ease of cross-model comparison, the sign of the X-axis coordinates for ViT-B has been inverted without altering the underlying conclusions. Generally, the average position of heads in the $i$-th layer lies between heads of the layers $i-1$ and $i+1$ for $i \in \{1, \dots, 10\}$. ViT-B demonstrates a comparable trend, though strong coordinate alignment is predominantly restricted to the initial and final layers. Although intermediate ViT-B heads show higher dispersion, partially due to the larger total head count, a smooth transitional flow remains visible across subsequent layers. 

Furthermore, the joint PCA across both model variants suggests that ViT-S samples the representation space more sparsely than ViT-B due to its reduced capacity, as shown in Figure~\ref{fig:commonPCA}b. Because the majority of ViT-B point neighborhoods contain ViT-S projections, both architectures appear to uncover similar semantic features, but ViT-B performs this with higher density and precision. When restricting the analysis to the top-24 heads selected by the exemplary \our{} models (Figure~\ref{fig:selected_heads_laplace}), this functional subset is scattered broadly across the space defined by two main principal components. The explained variance ratios for the first two components (derived from $150,000$ values) are $18.8\%$, $14.1\%$, and $14.0\%$ for ViT-S, ViT-B, and the joint model set, respectively. Appendix B.2 provides the detailed comparison of individual layers.

\begin{figure}[t]
    \centering
    \includegraphics{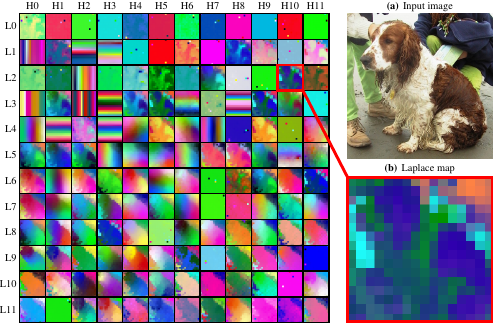}
    \caption{\textbf{DINOv2 head Laplacian visualization}. The (left) image displays the full Laplacian visualization map extracted from a DINOv2 ViT-B model over all layers and heads. On the (right) image a specific magnified region of the generated feature map (b) is compared alongside its corresponding input image (a).}
    \label{fig:semantic_clustering}
\end{figure}

\paragraph{Laplace maps}
Laplace maps discover the functional specialization of individual attention heads, enabling fine-grained, per-sample investigation. Across layers, consecutive heads attend to distinct visual hierarchies. As illustrated in Figure~\ref{fig:semantic_clustering}, individual heads isolate foreground-background boundaries, determine objects' shapes, or segment fine-grained semantic parts. Moreover, several heads exhibit functional redundancy by producing nearly identical spectral patterns, such as L4H1 and L4H3, or L6H0 and L6H2. Consistent with established findings on vision transformer representations~\cite{raghu2021vision,caron2021dino}, early layers predominantly act as local spatial and texture filters, which is evidenced by smooth positional gradients in L1H0 and L3H8 (which we term \emph{convolutional} heads). Deeper layers progressively shift toward more abstract semantics~\cite{raghu2021vision,caron2021dino}.

A distinct subset of heads exhibits highly localized eigenvector spectra characterized by a uniform background, with a few isolated, single-patch spikes (e.g. L0H1, L0H10, and L10H10). When this extreme mode concentration occurs consistently across samples, we designate the head as a \emph{router} head (analyzed in detail in Section~\ref{router_heads_analysis}). Rather than representing noise or pruning targets, we hypothesize that these heads serve as topological bottlenecks that dynamically aggregate and route contextual information across the attention graph.

Finally, analysis of the pruning patterns in Figures~\ref{fig:teaser} and~\ref{fig:selected_heads_laplace} reveals that \our{} preferentially retains deeper, semantically abstract heads, particularly in ViT-S. However, the most frequently preserved heads correspond to all three functional roles: \emph{convolutional} heads, \emph{router} heads, and heads identifying more complex shapes. This demonstrates that \our{} maintains a balanced structure rather than collapsing into a mono-semantic representation. Additional Laplace maps are provided in Appendix A.


\begin{table}[!b]
    \centering
    \footnotesize
    \caption{ 
   \textbf{FLOPs and parameters used by \our{}} and RAPTOR~\cite{jacobs2026raptor} in two ViT variants: ViT-S and ViT-B, for various numbers of heads $k$. Results denoted by ''{\color{black}\xmark}'' concern settings not designed for RAPTOR. Best results in each scenario are bolded when both RAPTOR and \our{} scores are calculated.}
    \begin{tblr}{
      colspec = {l c cccc},
      column{1} = {leftsep=0pt},
      vline{2,3,5} = {2-Z}{dashed},                  
      hline{2,4,Z} = {solid},
      hline{5,7,9,11,13} = {dashed},            
      rowsep = 1.5pt,
      colsep = 3.5pt,
      column{2} = {leftsep=6pt, rightsep=6pt},  
    }
    & & \SetCell[c=2]{c}\textbf{FLOPs} & & \SetCell[c=2]{c}\textbf{Parameters} & \\
    & & \textbf{ViT-S} & \textbf{ViT-B} & \textbf{ViT-S} & \textbf{ViT-B} \\
    \textbf{Method} & $\boldsymbol{k}$ & \textbf{\#GFLOPs} & \textbf{\#GFLOPs} & \textbf{\#Mparams} & \textbf{\#Mparams} \\
    DINOv2 & $-$ & $12.25$ & $46.33$ & $22.06$ & $86.58$ \\
    RAPTOR & 6 & {\color{black}\xmark}  & {\color{black}\xmark} & {\color{black}\xmark} & {\color{black}\xmark} \\
    SAPER & 6 & $7.80$ & $30.05$ & $15.56$ & $59.42$ \\
    RAPTOR & 12 & $16.15$ & {\color{black}\xmark} & $\mathbf{5.53}$  & {\color{black}\xmark} \\
    SAPER & 12 & $\mathbf{8.20}$ & $30.75$ & $16.15$ & $60.60$ \\
    RAPTOR & 24 & $16.15$ & $61.86$ & $\mathbf{7.91}$ & $\mathbf{20.49}$ \\
    SAPER & 24 & $\mathbf{9.01}$ & $\mathbf{32.17}$ & $17.33$ & $62.96$ \\
    RAPTOR & 36 & $16.15$ & $61.86$ & $\mathbf{10.30}$ & $\mathbf{29.98}$ \\
    SAPER & 36 & $\mathbf{9.82}$ & $\mathbf{33.58}$ & $18.51$ & $65.33$ \\
    RAPTOR & 48 & $16.15$ & $61.86$ & $\mathbf{12.68}$ & $\mathbf{39.47}$ \\
    SAPER & 48 & $\mathbf{10.63}$ & $\mathbf{35.00}$ & $19.69$ & $67.69$ \\
    \end{tblr}
    \label{tbl:saper_FLOPs}
\end{table}

\paragraph{Attention head selection}
In terms of computational cost, \our{} consistently achieves lower FLOPs than the base DINOv2 model across all head-selection budgets and both ViT backbones. Compared to RAPTOR, \our{} reduces GFLOPs by roughly a factor of two at matching head counts (e.g., $32.17$ vs. $61.86$ GFLOPs for ViT-B at $k=24$), 
despite retaining a higher parameter count than RAPTOR in the corresponding configurations (e.g, $62.96$M vs. $20.49$M parameters for ViT-B at $k=24$). This efficiency advantage stems from \our{}'s ability to select an arbitrary number of individual heads, rather than being restricted to RAPTOR's block-multiple constraint.

\begin{table*}[!ht]
    \centering
    \footnotesize
    \caption{\textbf{Overall performance of \our{}}, RAPTOR~\cite{jacobs2026raptor} and three head-selection baselines (random, magnitude and direct mask) in knowledge distillation mode on ImageNet-1K, for ViT-Small (ViT-S) and ViT-Base (ViT-B). RAPTOR accuracy is derived from Table 1 in~\cite{jacobs2026raptor}. Results of \our{} are averaged over 3 runs, while $k$ denotes the number of heads. The best result for each head allocation is bolded. Results denoted by ''{\color{black}\xmark}'' are not possible to obtain since the total number of heads used by RAPTOR must be a multiple of the number of heads in a single layer, and the authors do not consider single blocks.\label{tbl:saper_imagenet_distill}}
    \begin{tblr}{
    colspec = {l *{10}{c}},
    column{1} = {leftsep=0pt},
    vline{4,6,8,10} = {dashed},  
    vline{2} = {2-Z}{dashed},     
    rowsep = 1.5pt,
    colsep = 1.3pt,               
    column{1} = {rightsep=5pt},   
    column{Z} = {rightsep=0pt},
  }
  & \SetCell[c=10]{c} \textbf{ImageNet-1K (distilled)} \\
  \hline
  & \SetCell[c=2]{c} {$\boldsymbol{k = 6}$} &  & \SetCell[c=2]{c} {$\boldsymbol{k = 12}$} & & \SetCell[c=2]{c} {$\boldsymbol{k = 24}$} & & \SetCell[c=2]{c} $\boldsymbol{k = 36}$ & & \SetCell[c=2]{c} $\boldsymbol{k = 48}$ \\
  \textbf{Method} & \textbf{ViT-S} & \textbf{ViT-B} & \textbf{ViT-S} & \textbf{ViT-B} & \textbf{ViT-S} & \textbf{ViT-B} & \textbf{ViT-S} & \textbf{ViT-B} & \textbf{ViT-S} & \textbf{ViT-B} \\
  \hline
  RAPTOR & {\color{black}\xmark} & {\color{black}\xmark} & {\color{black}\xmark} & {\color{black}\xmark} & $-$ & $\mathbf{81.2_{\pm 0.2}}$ & $-$ & $\mathbf{83.0_{\pm 0.1}}$ & $-$ & $\mathbf{83.2_{\pm 0.1}}$ \\
  Random & $58.87_{\pm 3.43}$ & $62.49_{\pm 4.44}$ & $64.57_{\pm 2.38}$ & $68.32_{\pm 1.27}$ & $71.02_{\pm 0.88}$ & $72.65_{\pm 1.14}$ & $73.37_{\pm 1.02}$ & $75.50_{\pm 0.18}$ & $76.31_{\pm 0.91}$ & $77.35_{\pm 0.12}$ \\
  Magnitude & $56.41_{\pm 0.18}$ & $68.22_{\pm 0.20}$ & $66.19_{\pm 0.20}$ & $71.23_{\pm 0.07}$ & $70.61_{\pm 0.44}$ & $74.93_{\pm 0.32}$ & $74.63_{\pm 0.55}$ & $76.76_{\pm 0.25}$ & $78.39_{\pm 0.17}$ & $78.90_{\pm 0.11}$ \\
  Direct mask & $\mathbf{68.89_{\pm 4.16}}$ & $74.09_{\pm 0.69}$ & $74.39_{\pm 0.76}$ & $\mathbf{78.50_{\pm 0.23}}$ & $76.95_{\pm 0.38}$ & $80.41_{\pm 0.10}$ & $78.45_{\pm 0.12}$ & $80.89_{\pm 0.16}$ & $79.10_{\pm 0.16}$ & $81.71_{\pm 0.10}$ \\
  SAPER & $67.69_{\pm 1.74}$ & $\mathbf{75.42_{\pm 0.54}}$ & $\mathbf{74.68_{\pm 0.11}}$ & $75.48_{\pm 2.74}$ & $\mathbf{77.06_{\pm 0.14}}$ & $80.62_{\pm 0.57}$ & $\mathbf{78.55_{\pm 0.11}}$ & $81.67_{\pm 0.15}$ & $\mathbf{79.29_{\pm 0.15}}$ & $82.12_{\pm 0.31}$ \\
  \hline
    \end{tblr}
\end{table*}

RAPTOR was not evaluated for $k=6$ in the case of both ViT variants, and for $k=12$ with ViT-B, since its minimum head count is a multiple of the number of heads in a single layer (6 or 12 heads). Also, the authors of RAPTOR did not consider a single block for the entire network, so we excluded all such scenarios from the scores. At $k=24$, the two methods become directly comparable, and RAPTOR achieves a slightly higher accuracy than \our{}, while still requiring roughly twice the GFLOPs of \our{}. 



\our{} realizes an accuracy-efficiency trade-off, allowing one to choose any number of attention heads. Understandably, classification accuracy of \our{} is strongly dependent on the desired number of attention heads to be retained, often exceeding 10 p.p. between the results for 6 and 48 heads. What is particularly important, for 6 heads (out of 144), \our{} with ViT-B achieved over 75\% of classification accuracy, being only 5 p.p. worse compared to the model with 24 heads. Therefore, \our{} offers a massive reduction in the number of attention heads without catastrophic loss of accuracy and can be used for an in-depth analysis of the model behavior. Among the baselines, the direct mask achieved the highest performance, only slightly worse than \our{} and even better in some settings (e.g. ViT-B with 12 heads). The remaining methods, i.e. random or magnitude-based head selection, were much worse in all settings. Understandably, the higher the number of retained heads, the smaller the differences between the considered methods.

In CIFAR-100, \our{} consistently outperforms RAPTOR in every head budget for which a comparison is possible, all while using substantially fewer GFLOPs than RAPTOR. The results for CIFAR-100, as well as for ImageNet-1K in fine-tuning mode, and all details regarding the considered hyperparameters are described in Appendices F-G. In all experiments, excluding the fine-tuning of ViT-S on CIFAR-100, all attention heads were frozen throughout all training phases. Only the remaining network parameters were modified, since the adjustment of pre-trained attention heads was usually harmful to models.




\subsection{Router head identification}
\label{router_heads_analysis}
\begin{figure*}[t]
    \centering
    \includegraphics{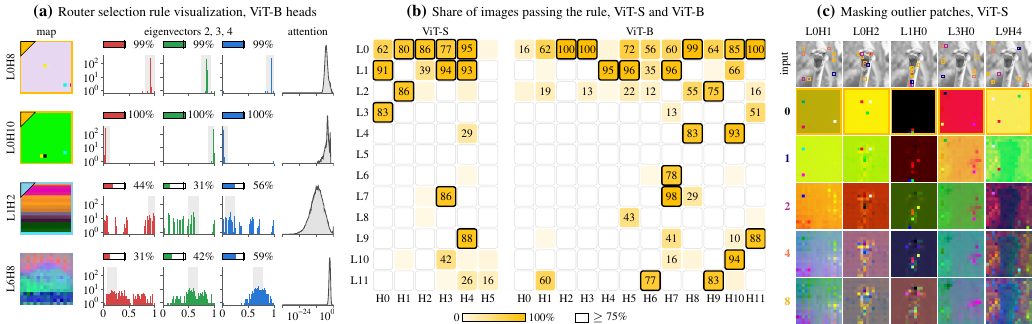}
    \caption{\textbf{\emph{Router} heads in DINOv2 without registers. }\textbf{(a)} The Laplace maps and histograms of eigenvectors 2 to 4 over the 256 patches with a $\pm0.1$ window around the mode (gray interval) in ViT-B. The bars above present the share of patches inside this interval. For comparison, the density of the attention weights is shown. \textbf{(b)} Share (\%) of the 50{,}000 ImageNet-1K validation images on which each head passes the \emph{router} rule in at least $75\%$ of the cases. \textbf{(c)} Laplace maps of five ViT-S \emph{router} heads without masking (0, framed) and after masking 1, 2, 4 and 8 outlying patches. The \emph{input} row marks the masked patches.}
    \label{fig:sink_heads}\label{fig:register_vit_s}\label{fig:register_heads_statistics}\label{fig:eigvec_channels}
\end{figure*}

\begin{table}[!b]
    \centering
    \footnotesize
    \caption{
    \textbf{Classification accuracy (\%) of a linear probe} on one last-layer token per image (CLS or a single patch token) for DINOv2 without registers. Mean and Std. over 5 seeds. Probes are trained on 64{,}000 ImageNet-1K training images and evaluated on the 50{,}000 validation images. Rows below the image count use only images where a last-layer head passes the \emph{router} rule. \emph{Router patch} is the outlying patch in that head's Laplace map, \emph{uniform patch} is a patch from the uniform region of the same map, \emph{max-norm patch} is the patch with the largest output norm~\cite{darcet2024registers}.
    }
    \begin{tblr}{
      width = \linewidth,
      colspec = {X[l] cc},
      column{1} = {leftsep=0pt},
      vline{2,3} = {2-Z}{dashed},                    
      hline{2,3,Z} = {solid},                   
      rowsep = 1.5pt,
      colsep = 6pt,
    }
    \textbf{Token} & \textbf{ViT-S} & \textbf{ViT-B} \\
    Random patch, all 50{,}000 images & $43.64_{\pm 0.14}$ & $47.78_{\pm 0.22}$ \\
    \SetCell[c=1]{l} \emph{Images with a last-layer router head} & \emph{15{,}874} & \emph{49{,}022} \\
    CLS & $71.10_{\pm 0.09}$ & $77.98_{\pm 0.03}$ \\
    Router patch & $37.70_{\pm 0.17}$ & $54.77_{\pm 0.08}$ \\
    Uniform patch, same head & $41.21_{\pm 0.31}$ & $47.69_{\pm 0.07}$ \\
    Max-norm patch & $43.83_{\pm 0.09}$ & $71.70_{\pm 0.02}$ \\
    \end{tblr}
    \label{tbl:token_probe}
\end{table}
  
During the forward pass of an image, we isolate \emph{router} heads using a spectral concentration criterion, identifying heads whose Laplace maps exhibit a uniform low-magnitude eigenvector mass paired with sparse outlier spikes. For each head, we independently normalize the three primary Laplace map channels to $[0, 1]$ across all $256$ spatial patches and estimate the distribution mode $S$ as the center of the most populated bin in a $100$-bin histogram. A head meets this per-image criterion if at least $96\%$ of its patch values fall within $[S - 0.1, S + 0.1]$.

As demonstrated in Figure~\ref{fig:eigvec_channels}, relying solely on the activation distribution is insufficient to isolate the \emph{router} heads, as defined above. Some attention heads exhibit similar distributions without actually targeting spatial outliers, indicating that distribution metrics must be paired with spatial or structural analysis. When we mask out outlying patches, as in Figure~\ref{fig:register_vit_s}, the underlying object shape emerges, revealing structural patterns that become increasingly complex in deeper layers. Also, for selected heads like L0H1, L3H0 or L9H4, the outlying patches are completely outside the main image object (here, an ostrich).

We evaluate the stability and distribution of \emph{router} heads across ViT architectures using the ImageNet-1K validation set and present the corresponding results in Figure~\ref{fig:register_heads_statistics}. In both ViT-S and ViT-B, the majority of \emph{router} heads are localized within the first two layers. However, specific deeper heads in ViT-B (e.g., L7H7, L9H11, L10H10, and L11H9) consistently satisfy our criteria across the vast majority of samples (98\%, 88\%, 94\%, and 83\%, respectively). Similarly, in ViT-S, heads L7H3 and L9H4 exhibit strong \emph{router}-like behavior (86\% and 88\% of samples), whereas this prevalence drops significantly in the final two layers. In Appendix C, we present the same analysis for DINOv2 with registers. Interestingly, also in such a setting, \our{} isolates dedicated heads characterized by extreme eigenvector mode concentration, confirming that it remains a distinct structural role.

For selected pruned models with \our{}, we evaluated the individual head importance by measuring the accuracy drop after removing (zeroing) consecutive heads. As shown in Figure~\ref{fig:pruned_models}c, the presence of \emph{router} heads was important for the exemplary fine-tuned model and is more beneficial than in the distillation mode, as compared in Appendix E. In an extreme case, for one of the 12-head models, removing L8H5 and L11H6 led to even 53.6\% and 31.3\% accuracy drops, respectively.
    
Following~\cite{darcet2024registers}, we present in Table~\ref{tbl:token_probe} linear probing experiments based on the last layer representation. We compared classification accuracy using the pure CLS token, randomly selected spatial tokens, \emph{router tokens}, identified through Laplace maps of the last layer's attention heads, and high-norm outliers defined by~\cite{darcet2024registers}. The topological routers outperformed random spatial patches in ViT-B, but yielded lower accuracy than the aggregated high-norm outliers.
    
Our unnormalized Laplacian formulation reveals that DINOv2 relies on specialized routing heads characterized by extreme eigenvector mode concentration. While prior work demonstrates that ViTs actively repurpose patch tokens into implicit memory registers~\cite{darcet2024registers}, these artifacts have been identified through feature-norm magnitudes. Our spectral analysis suggests the structural mechanics driving this phenomenon: because their token indices shift dynamically with the input image, we hypothesize these coordinates function as data-driven \textit{router heads}, being active topological bottlenecks rather than static storage. Aligning with factorized attention mechanics~\cite{doshi2026biorthogonal}, these structural hubs act as physical bridges for content-position coupling. Single-token probes suggest that these router tokens store global context,  outperforming random spatial patches while disaggregating the fully accumulated feature-norm artifacts. Consequently, we characterize these structures as dynamic routing circuits that actively bridge spatial topology with attention-driven context aggregation. Therefore, \our{} may be used as a finer-grained interpretability mechanism. However, in this paper, we focused solely on presenting this phenomenon and outlining its characteristics. A precise explanation of this behavior requires a detailed analysis that goes beyond the scope of this work.

\section{Conclusions}
We introduce \our{}, a unified framework that places explainability at the center of DINOv2 analysis and compression. Through our graph-spectral visualization method, we uncover the fine-grained functional specialization of individual attention heads, distinguishing among \emph{convolutional} heads, \emph{router} heads, and heads capturing complex, global semantic structures. Moreover, we demonstrate that functionally analogous heads operate not only within contiguous blocks, but also across varying network depths. Guided by these structural insights, \our{} leverages LapSum for exact-budget, differentiable head selection, demonstrating competitive classification accuracy while substantially reducing FLOPs. Ultimately, this work advances transparent vision foundation models by providing principled spectral diagnostics that both explain internal attention mechanics and inform faithful, budget-constrained model reduction.

In future work, we plan to extend \our{} to fine-grained token-level sparsification, enforcing both per-head and global network-wide token budgets, as well as joint structural pruning of MLP layers.

\section*{LLM Usage Statement}
During the preparation of this publication, we utilized LLMs for writing experimental code, generating plots, consulting on specific ideas, conducting literature reviews, and editing or proofreading the text. All outputs generated by the LLMs were verified and validated by humans. The research was not conducted using autonomous agents.

{
    \small
    \bibliographystyle{ieeenat_fullname}
    \bibliography{main}
}

\clearpage
\setcounter{page}{1}

\onecolumn
\appendix   

{\centering
{\Large\textbf{Pruning as a Lens: Understanding the Functional Roles\\of Attention Heads in Vision Transformers}\par}
\vspace{0.4em}
{\large Supplementary Material\par}}


\section{Laplace maps}\label{app:laplace}
\begin{figure}[!ht]
  \centering
  \includegraphics[width=\linewidth]{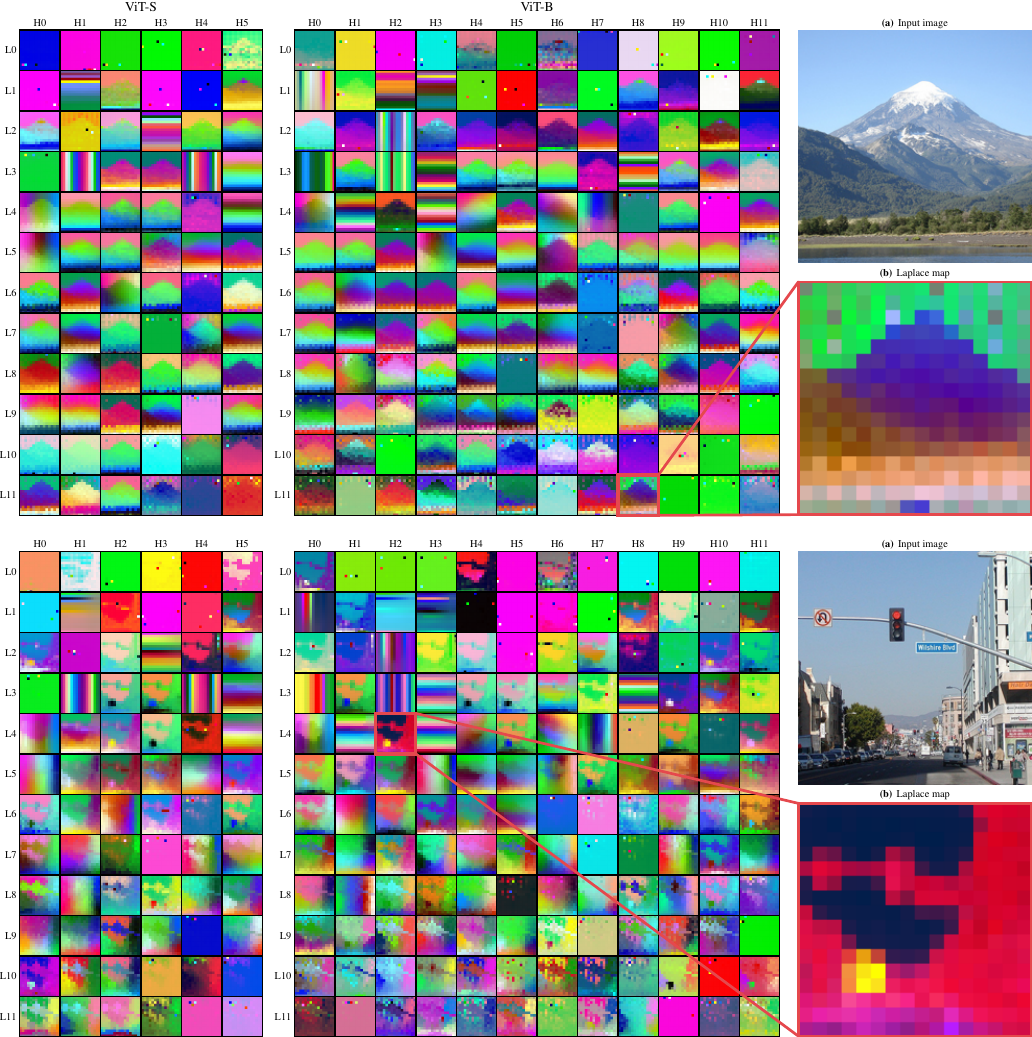}
  \caption{Laplace maps for the two selected samples of the ImageNet-1K dataset for all attention heads of ViT-S (left column) and ViT-B (middle column), with corresponding input images \textbf{(a)}, i.e. a mountain and a traffic light, and enlarged selected Laplace maps \textbf{(b)}.}\label{laplace_maps_full_1}
\end{figure}

\begin{figure}[!ht]
  \centering
  \includegraphics[width=\linewidth]{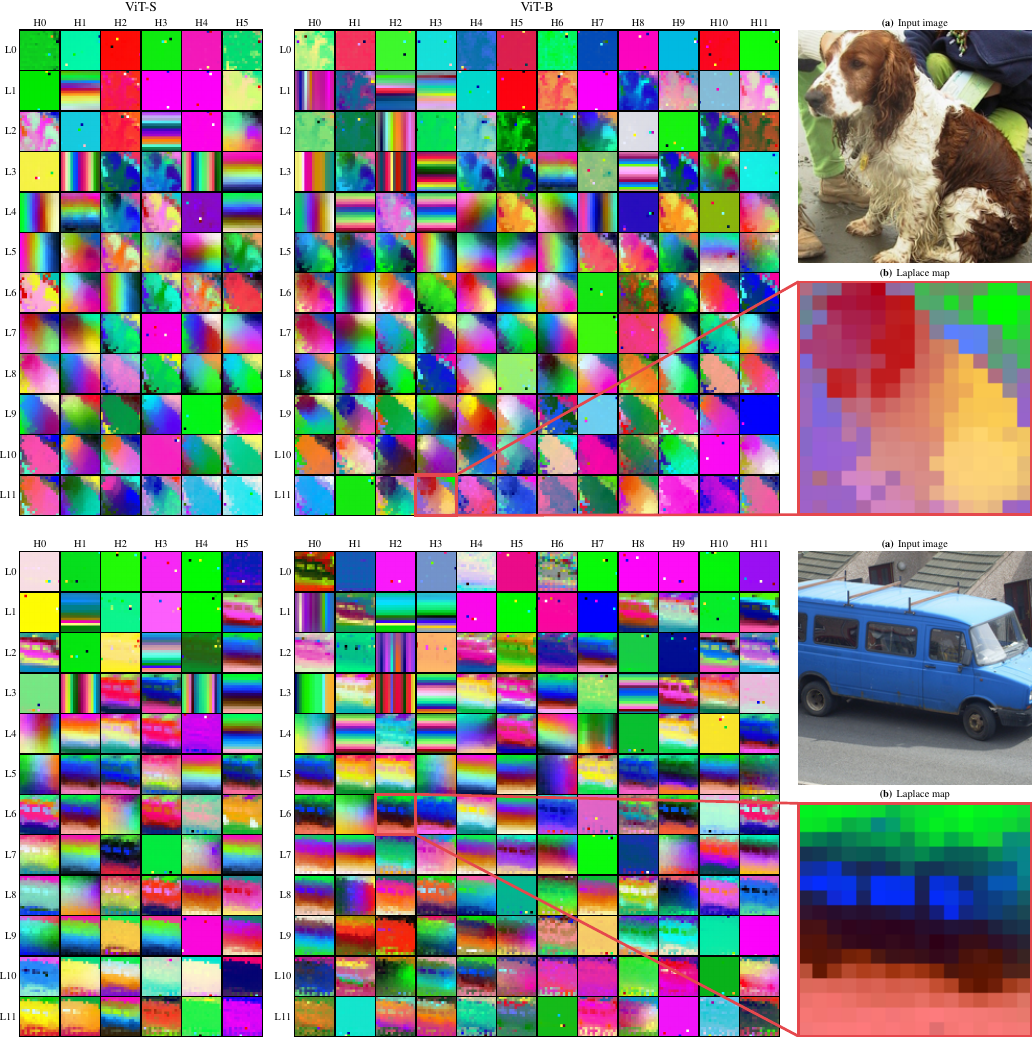}
  \caption{Laplace maps for the two selected samples of the ImageNet-1K dataset for all attention heads of ViT-S (left column) and ViT-B (middle column), with corresponding input images \textbf{(a)}, i.e. a dog and a car, and enlarged selected Laplace maps \textbf{(b)}.\label{laplace_maps_full_2}}
\end{figure}

\begin{figure}[!ht]
  \centering
  \includegraphics[width=\linewidth]{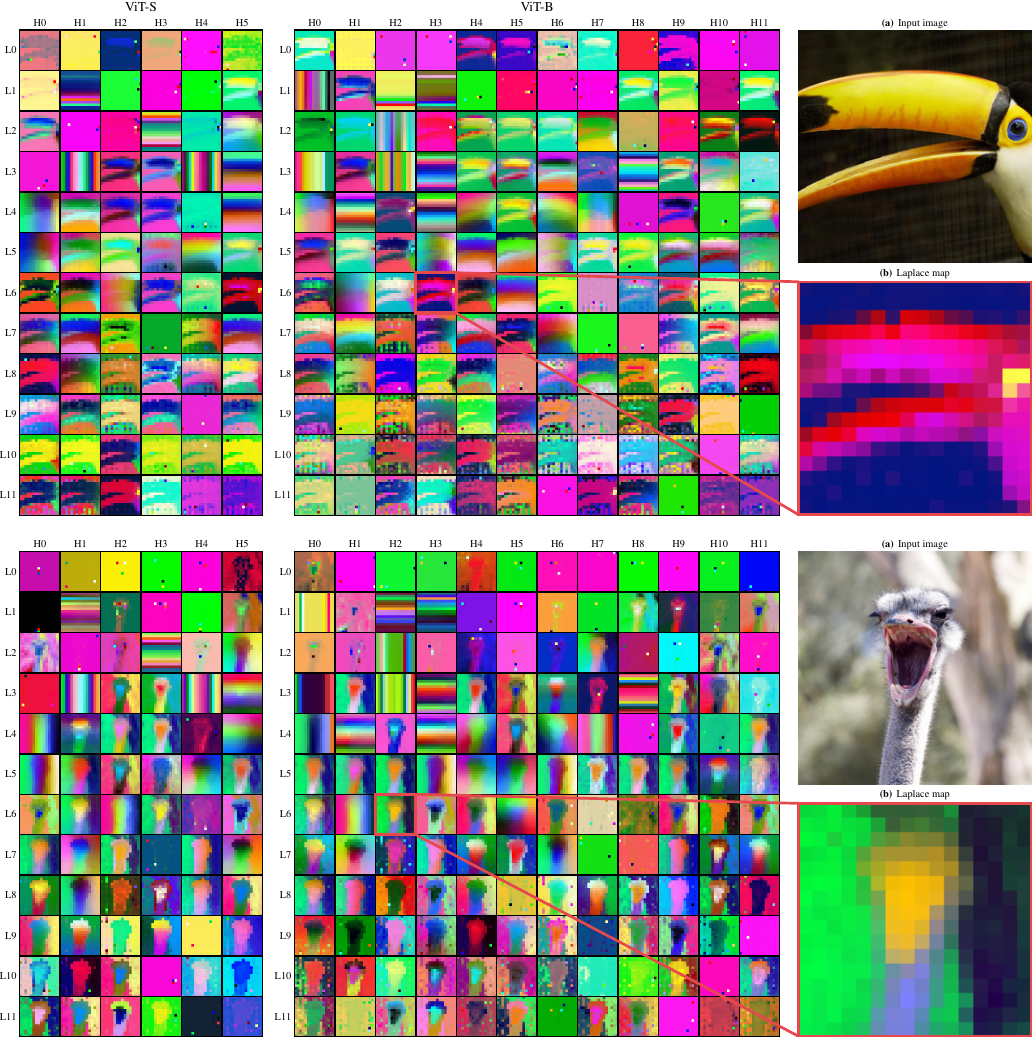}
  \caption{Laplace maps for two further samples of the ImageNet-1K dataset for all attention heads of ViT-S (left column) and ViT-B (middle column), with corresponding input images \textbf{(a)}, i.e. a toucan and an ostrich, and enlarged selected Laplace maps \textbf{(b)}.}\label{laplace_maps_full_3}
\end{figure}
\begin{figure}[!ht]
  \centering
  \includegraphics[width=\linewidth]{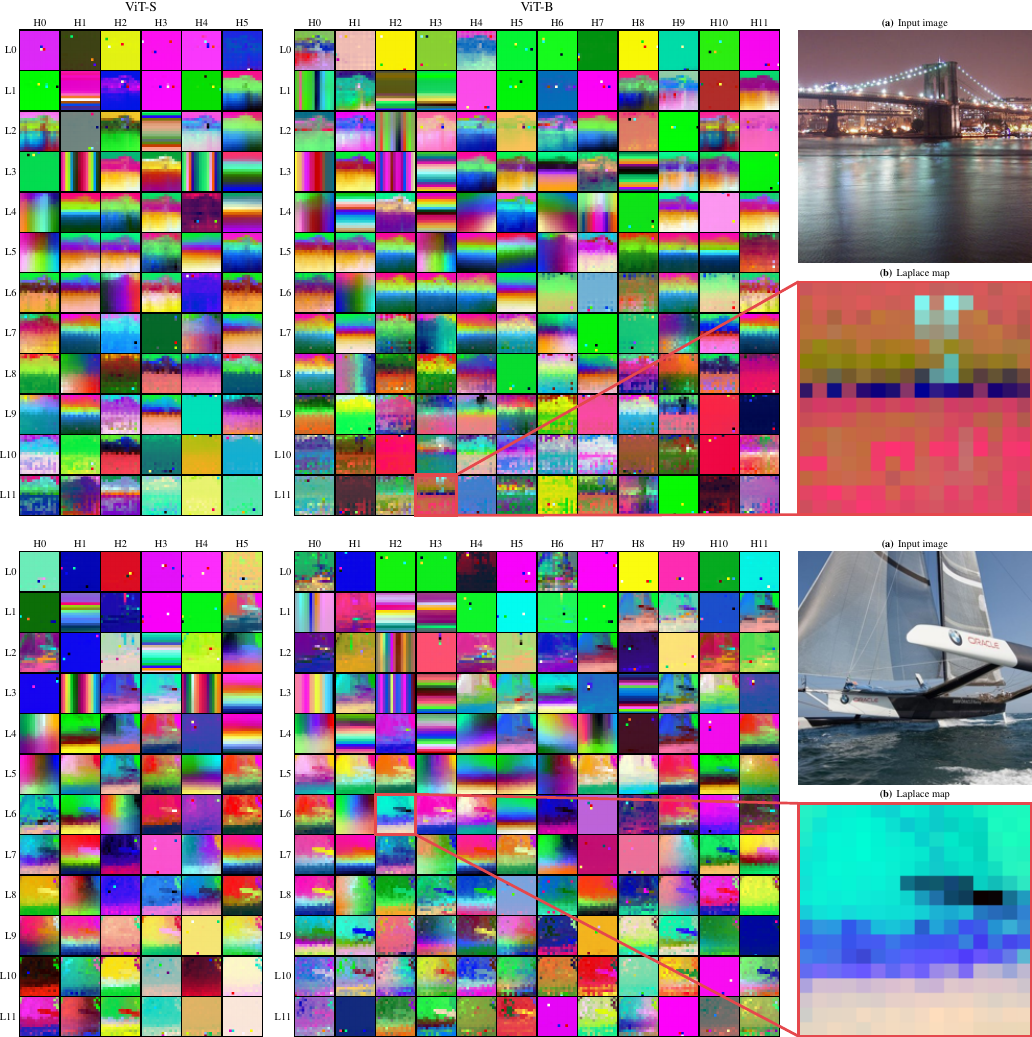}
  \caption{Laplace maps for two further samples of the ImageNet-1K dataset for all attention heads of ViT-S (left column) and ViT-B (middle column), with corresponding input images \textbf{(a)}, i.e. a pier and a trimaran, and enlarged selected Laplace maps \textbf{(b)}.}\label{laplace_maps_full_4}
\end{figure}

\begin{figure}[p]
    \centering
    \includegraphics[width=\linewidth]{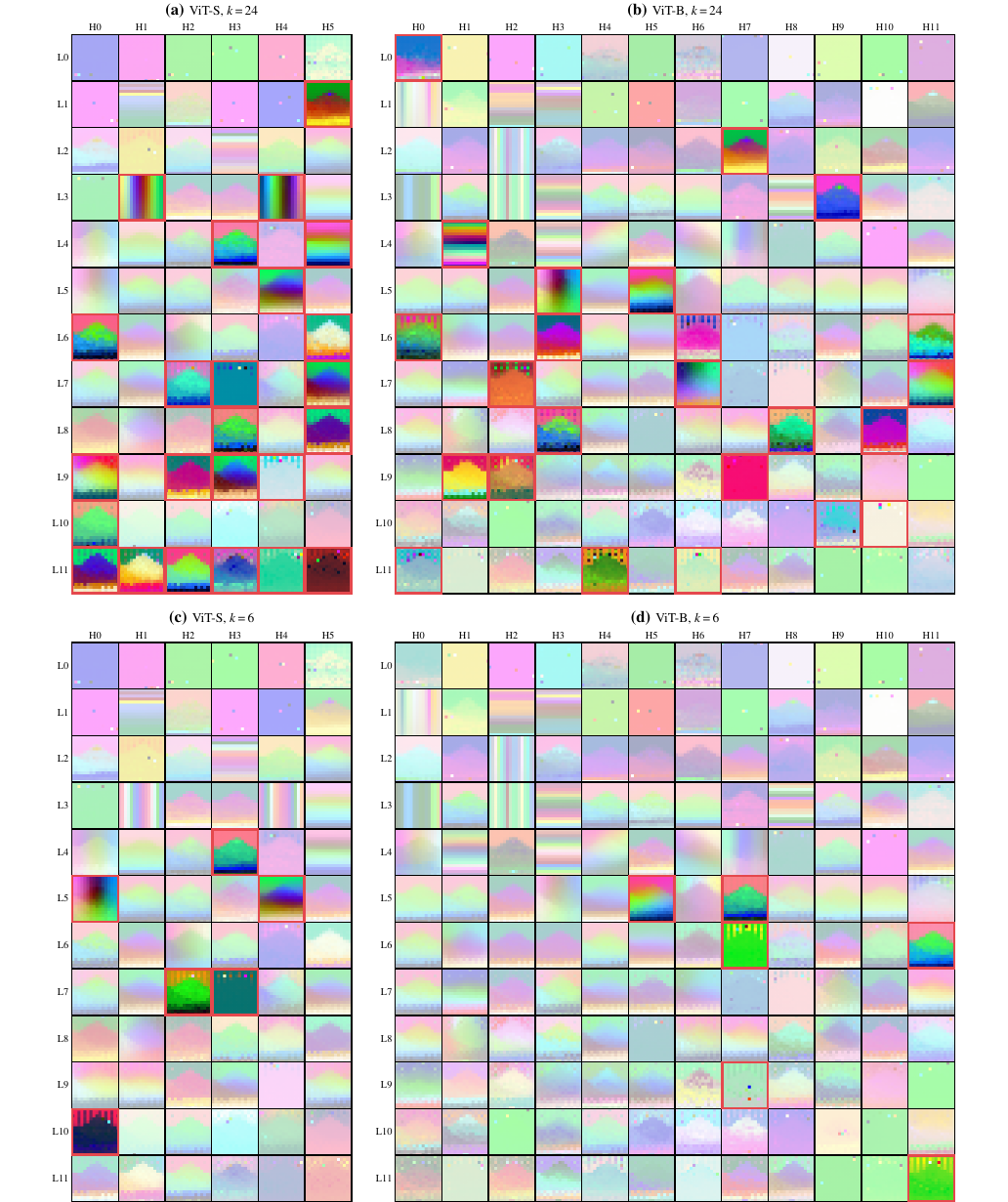}
    \caption{Laplace visualization maps for selected ViT-S and ViT-B models trained on the ImageNet-1K dataset, for an exemplary sample. Semi-transparent heads were selected for pruning by \our{}, while fully visible ones, bordered in red, were among the Top-24 (\textbf{a}, \textbf{b}) or Top-6 (\textbf{c}, \textbf{d}) heads.}
    \label{fig:selected_heads_laplace_supplement}
\end{figure}

\begin{figure*}[p]
  \centering
  \includegraphics[width=\textwidth]{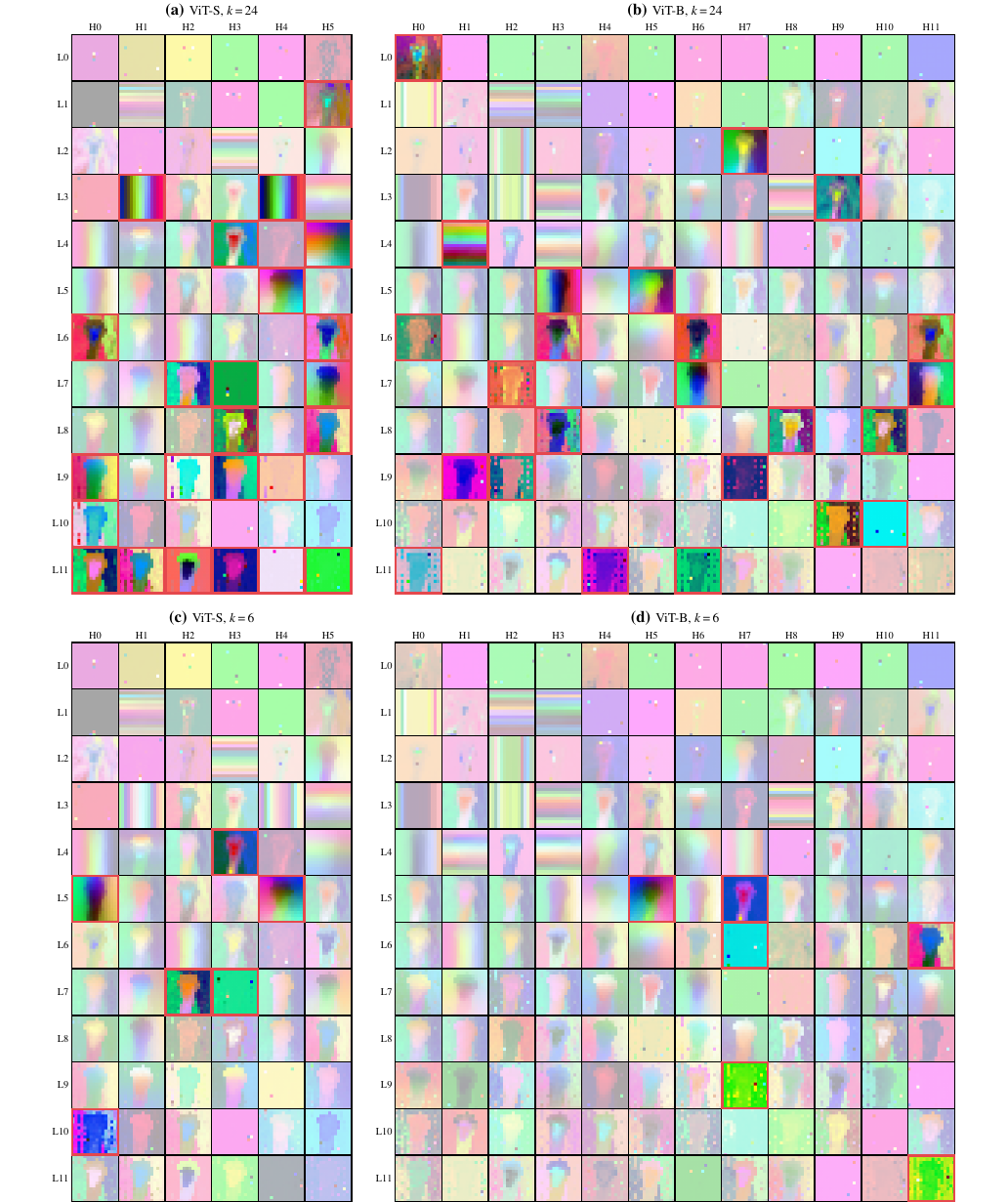}
  \caption{Continuation of Figure~\ref{fig:selected_heads_laplace_supplement} on the ostrich (input and full model: Figure~\ref{laplace_maps_full_3}) for the same four pruned models (distillation; $k = 24$ in (\textbf{a}, \textbf{b}), $k = 6$ in (\textbf{c}, \textbf{d})), where retained heads are bordered in red, and pruned heads are semi-transparent.}
  \label{fig:app_pruned_maps_ostrich}
\end{figure*}
\begin{figure*}[p]
  \centering
  \includegraphics[width=\textwidth]{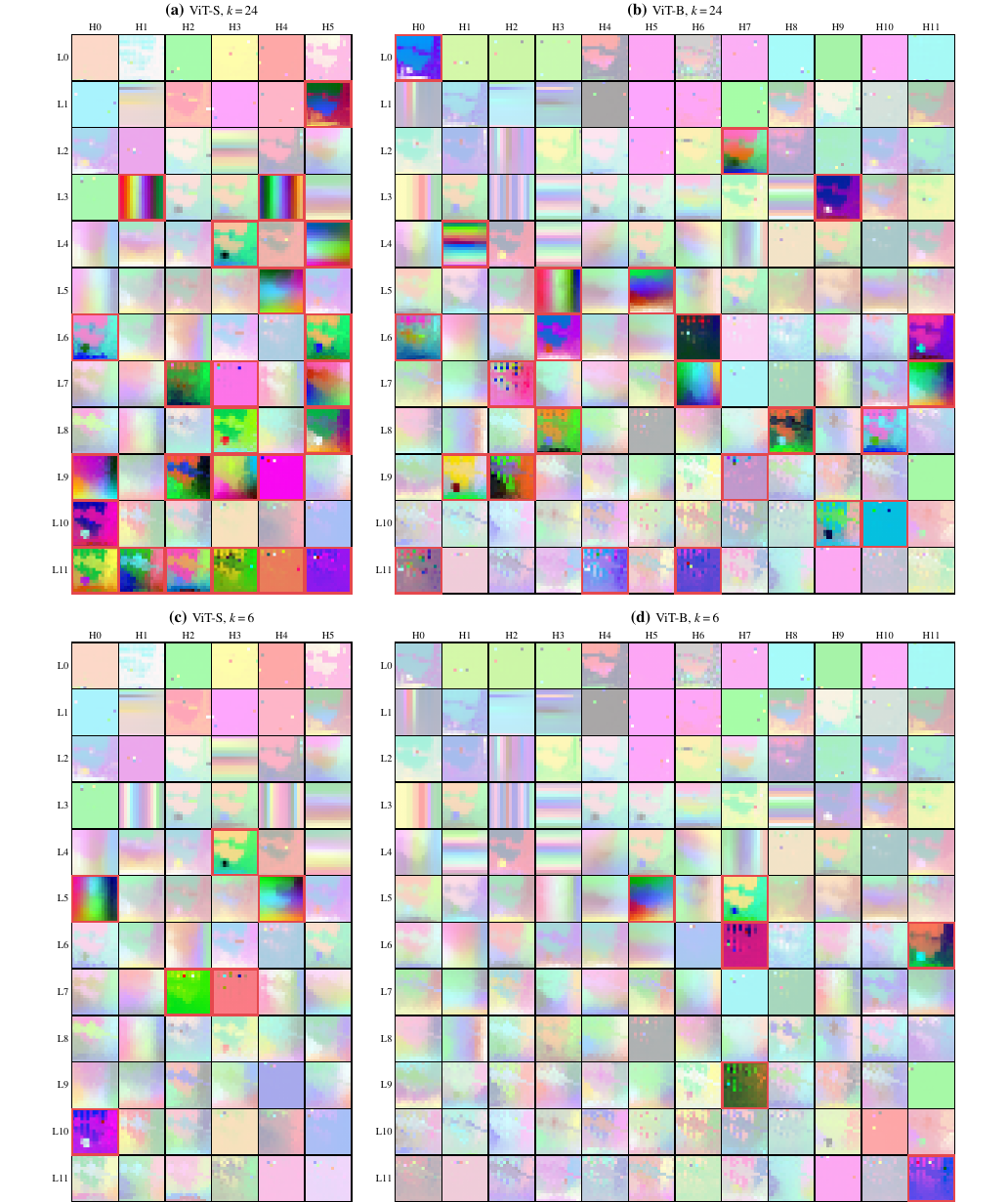}
  \caption{Continuation of Figures~\ref{fig:selected_heads_laplace_supplement}--\ref{fig:app_pruned_maps_ostrich}, on the traffic light image (input and full model: Figure~\ref{laplace_maps_full_1}) for the same four pruned models (distillation; $k = 24$ in (\textbf{a}, \textbf{b}), $k = 6$ in (\textbf{c}, \textbf{d})), where retained heads are bordered in red, and pruned heads are semi-transparent.}
  \label{fig:app_pruned_maps_traffic_light}
\end{figure*}

Figures~\ref{laplace_maps_full_1}--\ref{laplace_maps_full_4} present Laplace maps for all attention heads (and layers) for both considered ViT variants and eight selected images from the ImageNet-1K dataset. They demonstrate the diversity of tasks of individual heads (detecting edges, simple or more abstract shapes, etc.). Also, while in many cases their roles are rather clear, for some heads they are more subtle (like L11H10 in ViT-B). In the case of early-layer heads, for the presented images, the corresponding Laplace maps are almost single-colored, excluding individual pixels, demonstrating the pattern of \emph{router} heads. Additionally, Figures~\ref{fig:selected_heads_laplace_supplement}--\ref{fig:app_pruned_maps_traffic_light} depict Laplace maps selected by exemplary \our{} models during pruning to 24 and 6 heads, for the same image as shown in the first row of Figure~\ref{laplace_maps_full_1}. It may be concluded that \our{} retained heads responsible for various tasks, including edge detection, but also \emph{router} heads.

Figures~\ref{fig:app_pruned_union_vit_s} and~\ref{fig:app_pruned_union_vit_b} show Laplace maps of every head retained at $k=6$ or $k=12$ in distillation or fine-tuning (seed 42; 24 heads in ViT-S, 27 in ViT-B) on nine images, for the full model and a pruned one. Most heads keep similar maps: the median subspace overlap of their eigenvectors 2--4 is 0.73 in ViT-S and 0.59 in ViT-B. The lowest overlap values belong to \textit{router} heads (ViT-S L7H3 and L9H4, ViT-B L10H10), whose flat maps leave the eigenvectors undetermined, and to late-layer heads (ViT-S L11H4 and L11H5; ViT-B L8H5, L9H7 and L10H9). 

\newpage
Figures~\ref{fig:app_eigvec_ten_vit_s} and~\ref{fig:app_eigvec_ten_vit_b} present the ablation study described in Section~3.2, i.e. the input image and the first ten smallest eigenvectors of selected representative heads of each model. Eigenvectors beyond the fourth smallest keep the character of the head and split the image into finer parts, while every eigenvector of a \textit{router} head concentrates on one or two patches.
\begin{figure*}[p]
  \centering
  \includegraphics[scale=0.98]{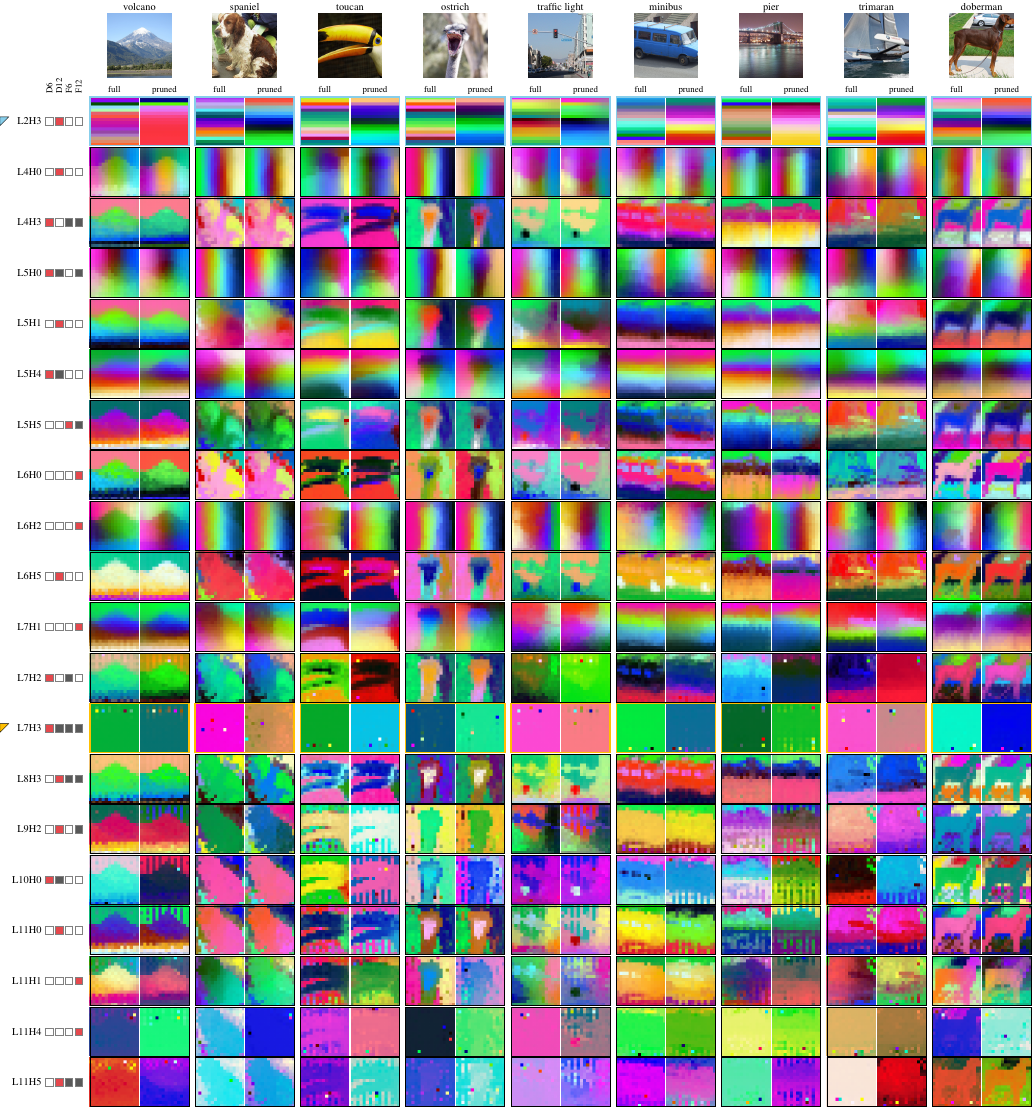}
  \caption{Laplace maps before and after pruning for each retained ViT-S head at $k=6$ or $k=12$ in distillation or fine-tuning with a single seed and 24 heads, shown in the eight images depicted in Figures~\ref{laplace_maps_full_1}--\ref{laplace_maps_full_4} and one more ImageNet-1K validation image. In each pair of columns, i.e. the head in the full DINOv2 (left) and in a pruned model (right), the pruned model's eigenvectors are sign-aligned to the full model's. Squares on the left side of the maps correspond to the runs that keep the head (D: distillation, F: fine-tuning; the number is $k$). The pruned map comes from the run marked in red, the first of D6, D12, F6, F12 that retains the head. Dark gray squares mark the other runs that also retain the head, but their maps are not shown. \emph{Router} heads are marked by the amber triangle, while no \textit{convolutional} head is retained at the considered $k$ values.}
  \label{fig:app_pruned_union_vit_s}
\end{figure*}
\begin{figure*}[p]
  \centering
  \includegraphics{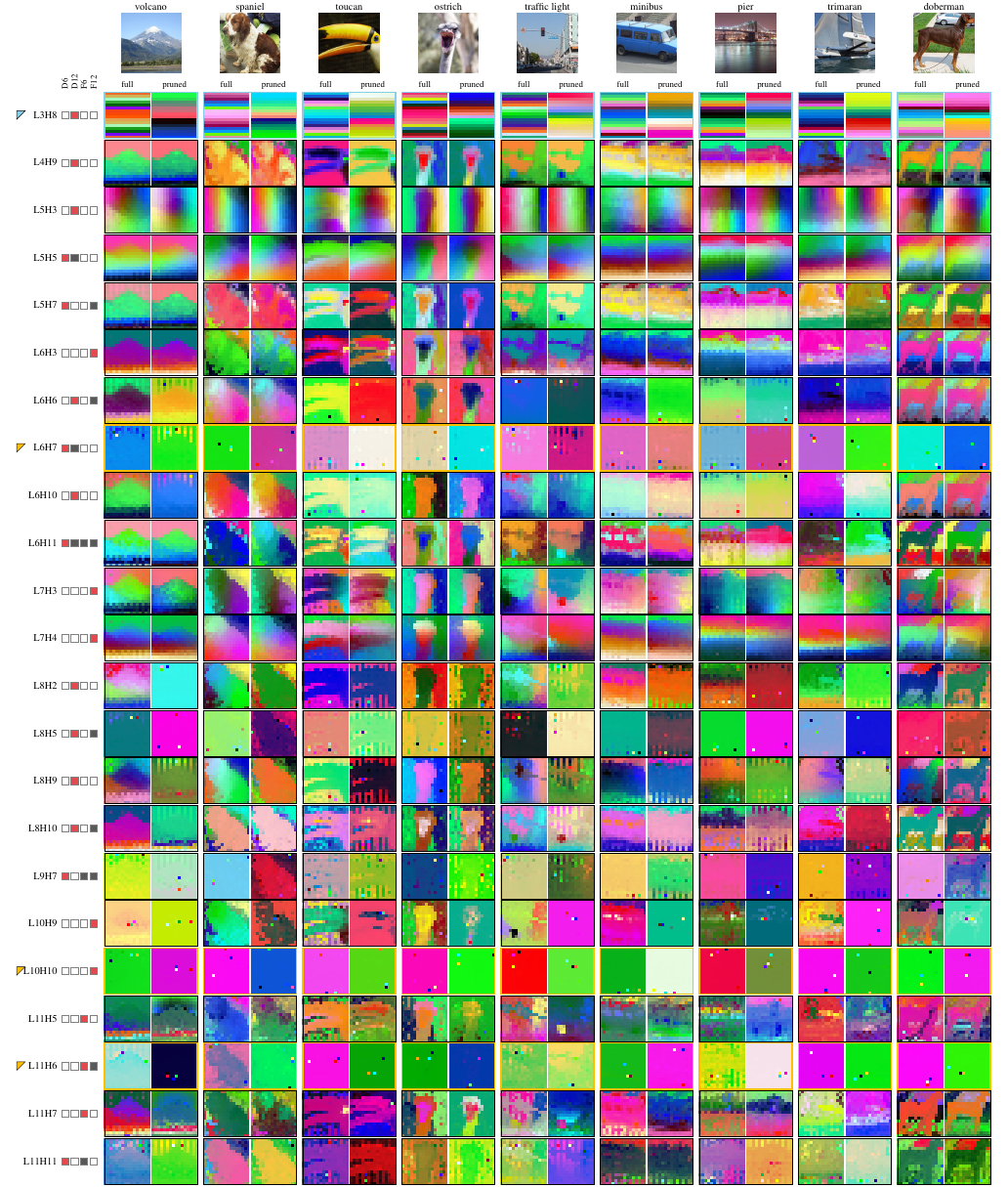}
  \caption{Continuation of Figure~\ref{fig:app_pruned_union_vit_s} for 27 heads from ViT-B.}
  \label{fig:app_pruned_union_vit_b}
\end{figure*}

\begin{figure*}[p]
  \centering
  \includegraphics{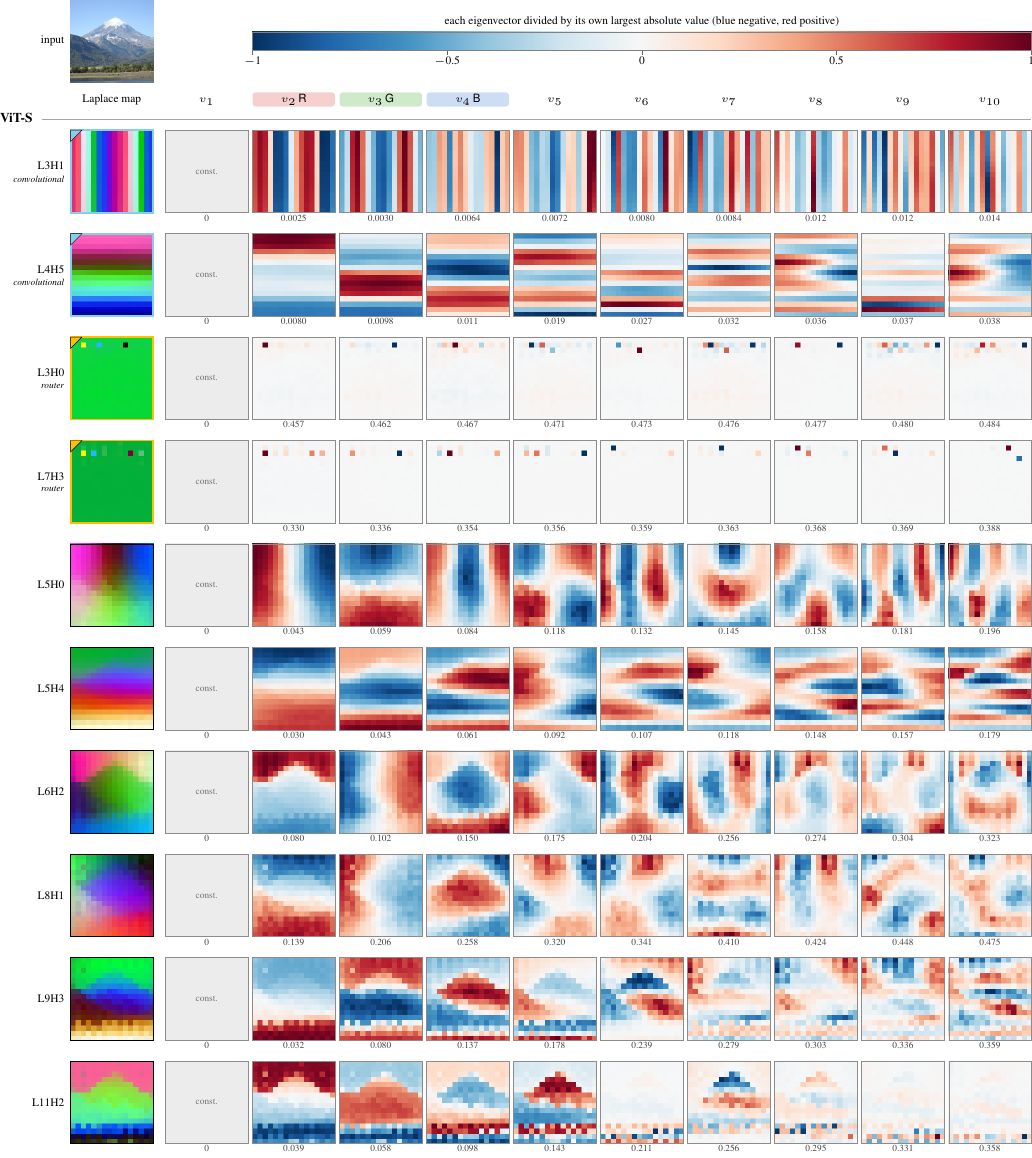}
  \caption{The first ten Laplacian eigenvectors $v_1,\dots,v_{10}$ of ten representative heads of ViT-S on the volcano presented in Figure~\ref{laplace_maps_full_1}, being the ablation of Section~3.2 for two \textit{convolutional} heads (vertical and horizontal), two \textit{router} heads (an early and a late one) and six other heads from layer 5 to the last. The left column presents the Laplace map ($v_2$, $v_3$, $v_4$ as R, G, B), and the next columns correspond to consecutive eigenvectors in a diverging scale (blue negative, red positive) scaled to their largest absolute value, while $v_1$ remains constant. Under each eigenvector, its eigenvalue is described. Higher eigenvectors keep the character of the head, i.e. stripes along one axis for the \textit{convolutional} heads, and 2-D abstract shapes. Every eigenvector of a \textit{router} head concentrates on a few patches (median 2 of 256, against 111 for the other heads shown). None of the other heads shown looks like a \textit{router} on any of the eight images.}
  \label{fig:app_eigvec_ten_vit_s}
\end{figure*}
\begin{figure*}[p]
  \centering
  \includegraphics{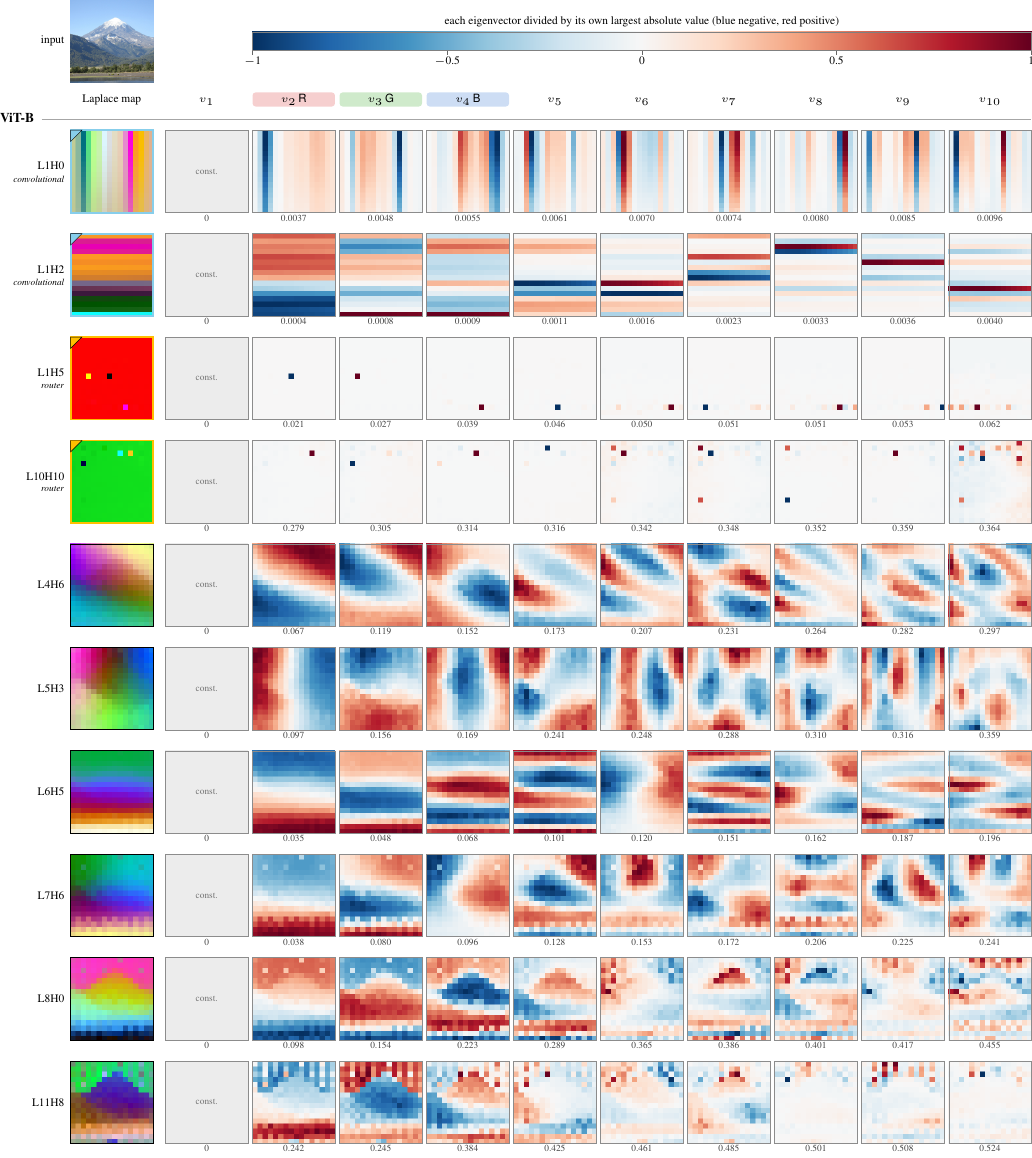}
  \caption{Continuation of Figure~\ref{fig:app_eigvec_ten_vit_s} for ViT-B: two \textit{convolutional} heads, two \textit{router} heads and six other heads from layer 4 to the last. Every eigenvector of a \textit{router} head concentrates on a few patches (median 1 of 256, against 95 for the other heads shown).}
  \label{fig:app_eigvec_ten_vit_b}
\end{figure*}

\clearpage
\section{Semantic clustering and PCA}\label{app:structure}
\begin{figure}[!h]
  \centering
  \begin{subfigure}[b]{171.28pt}\includegraphics{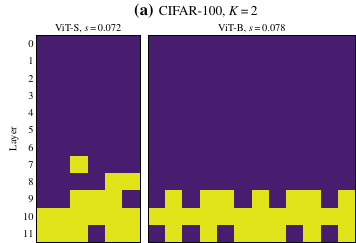}\phantomsubcaption\label{fig:cifar_vits_two_clusters}\label{fig:cifar_vitb_2_cluster}\end{subfigure}%
  \hspace{9.0pt}%
  \begin{subfigure}[b]{153.78pt}\includegraphics{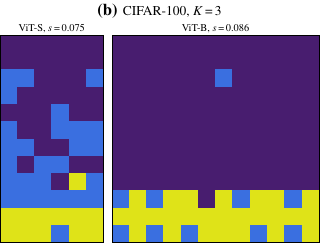}\phantomsubcaption\label{fig:cifar_vits_three_clusters}\label{fig:cifar_vitb_3_cluster}\end{subfigure}%
  \hspace{9.0pt}%
  \begin{subfigure}[b]{153.78pt}\includegraphics{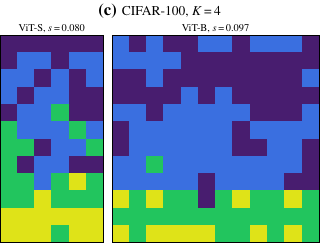}\phantomsubcaption\label{fig:cifar_vits_four_clusters}\label{fig:cifar_vitb_4_cluster}\end{subfigure}%
  \par\vspace{6.0pt}
  \begin{subfigure}[b]{171.28pt}\includegraphics{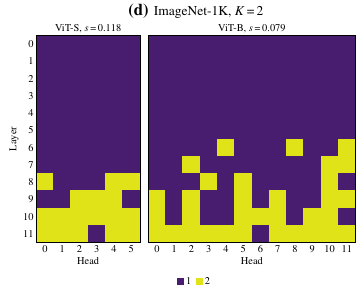}\phantomsubcaption\label{fig:imagenet_vits_two_clusters}\label{fig:imagenet_vitb_2_cluster}\end{subfigure}%
  \hspace{9.0pt}%
  \begin{subfigure}[b]{153.78pt}\includegraphics{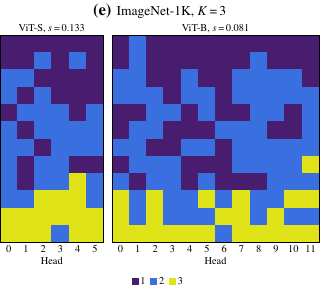}\phantomsubcaption\label{fig:imagenet_vits_three_clusters}\label{fig:imagenet_vitb_3_cluster}\end{subfigure}%
  \hspace{9.0pt}%
  \begin{subfigure}[b]{153.78pt}\includegraphics{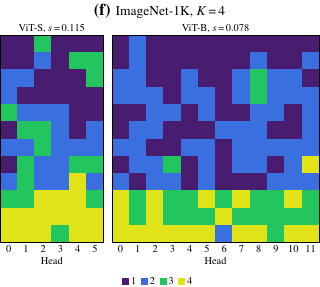}\phantomsubcaption\label{fig:imagenet_vits_four_clusters}\label{fig:imagenet_vitb_4_cluster}\end{subfigure}%
  \caption{Semantic clustering of the attention heads of ViT-S and ViT-B on CIFAR-100 (top row) and ImageNet-1K (bottom row) into $K = 2, 3$ and $4$ clusters, numbered by layers and heads, while $s$ is the silhouette score. In both models, the last two or three layers separate from the rest, in line with RAPTOR. However, our results paint a more nuanced picture, not fitting into a rigid layers-only division.}
  \label{fig:app_clustering}\label{fig:vits_cifar100_clustering}\label{fig:vits_imagenet_clustering}\label{fig:vitb_cifar100_clustering}\label{fig:vitb_imagenet_clustering}
\end{figure}

\subsection{Semantic clustering}\label{app:semantic}
\begin{table*}[!ht]
    \centering
    \footnotesize
    \begin{tblr}{
      colspec = {l l l *{9}{c}},
      column{1} = {leftsep=0pt},
      column{Z} = {rightsep=0pt},
      vline{2,4} = {1-Z}{dashed},
      hline{2,Z} = {solid},
      hline{6} = {solid},
      hline{4,8} = {dashed},
      rowsep = 1.2pt,
      colsep = 4pt,
    }
    \textbf{Measure} & \textbf{Dataset} & \textbf{Model} & $K=2$ & $K=3$ & $K=4$ & $K=5$ & $K=6$ & $K=7$ & $K=8$ & $K=9$ & $K=10$ \\
    \SetCell[r=4]{l} Silhouette $\uparrow$ & ImageNet-1K & ViT-S & 0.118 & 0.133 & 0.115 & 0.132 & 0.127 & 0.129 & \textbf{0.140} & 0.116 & 0.127 \\
     & ImageNet-1K & ViT-B & \textbf{0.082} & 0.081 & 0.078 & 0.081 & 0.079 & 0.067 & 0.081 & 0.075 & 0.074 \\
     & CIFAR-100 & ViT-S & 0.072 & 0.075 & 0.080 & 0.071 & 0.088 & \textbf{0.103} & \textbf{0.103} & 0.099 & 0.091 \\
     & CIFAR-100 & ViT-B & 0.078 & 0.086 & 0.097 & 0.102 & 0.099 & 0.094 & 0.073 & \textbf{0.109} & 0.104 \\
    \SetCell[r=4]{l} Davies--Bouldin $\downarrow$ & ImageNet-1K & ViT-S & \textbf{1.109} & 1.507 & 2.267 & 1.630 & 1.897 & 1.982 & 1.739 & 1.842 & 1.843 \\
     & ImageNet-1K & ViT-B & \textbf{1.372} & 1.730 & 2.012 & 2.299 & 2.171 & 2.149 & 2.262 & 2.402 & 2.112 \\
     & CIFAR-100 & ViT-S & 2.055 & 1.848 & 1.972 & 2.009 & 1.963 & 1.805 & 1.785 & \textbf{1.731} & 1.947 \\
     & CIFAR-100 & ViT-B & \textbf{1.328} & 1.377 & 2.022 & 1.876 & 1.943 & 2.003 & 2.083 & 1.964 & 2.077 \\
    \end{tblr}
    \caption{Evaluation of the spectral clustering of the heads (Appendix~\ref{app:semantic}) for $K = 2$ to $10$ clusters with silhouette and Davies--Bouldin index on the correlation of the heads' peak CLS attention over the ImageNet-1K validation set (50{,}000 images) and the CIFAR-100 test set (10{,}000). The best $K$ of each row is in bold.}
    \label{tbl:app_clustering_quality}
\end{table*}

In ViT-S, nearly all heads from the final two layers form a distinct cluster, which also incorporates selected heads from L8--L9, depending on the specific clustering and dataset configuration. This pattern indicates the emergence of highly specialized representations within the final three to four layers, particularly in L10 and L11. When the cluster granularity is increased, subsequent substructures are formed primarily from cluster 1 of the two-cluster setting (see Figures~\ref{fig:cifar_vits_two_clusters}~and~\ref{fig:imagenet_vits_two_clusters}). Crucially, these assignments reveal that individual heads at varying network depths frequently produce similar activations. This nuanced behavior is obscured by the layer-level block split performed by RAPTOR.

ViT-B exhibits comparable structural dynamics, while the strong substructure is notable in the final three layers. In CIFAR-100, no heads from L0--L8 belong to cluster 1, while for ImageNet-1K, isolated heads from L6--L8 belong to the cluster dominated by the last layer heads (compare Figures~\ref{fig:cifar_vitb_2_cluster} and \ref{fig:imagenet_vitb_2_cluster}). While a four-cluster setup consistently partitions clusters 1 and 2 from the two-cluster setup (see Figures~\ref{fig:cifar_vitb_4_cluster} and \ref{fig:imagenet_vitb_4_cluster}) into further substructures, a three-cluster configuration yields dataset-dependent behaviors. Specifically, the third cluster in CIFAR-100 isolates a subset of heads from the final three ViT-B layers, while for ImageNet-1K it contains individual heads scattered throughout the entire network. This indicates that the macro-level observations, such as distinct last layer representations, generalize across datasets, while fine-grained head assignments remain dataset-specific.

The silhouettes (Table~\ref{tbl:app_clustering_quality}) stay between 0.07 and 0.14 for 2 to 10 clusters, for both models and both datasets. The clusters are weakly separated, and no number of clusters stands out drastically, so we interpret them as tendencies of the heads rather than sharp and concise groups.

\subsection{Principal component analysis}\label{app:pca}
\begin{figure}[!h]
  \centering
  \includegraphics[width=\linewidth]{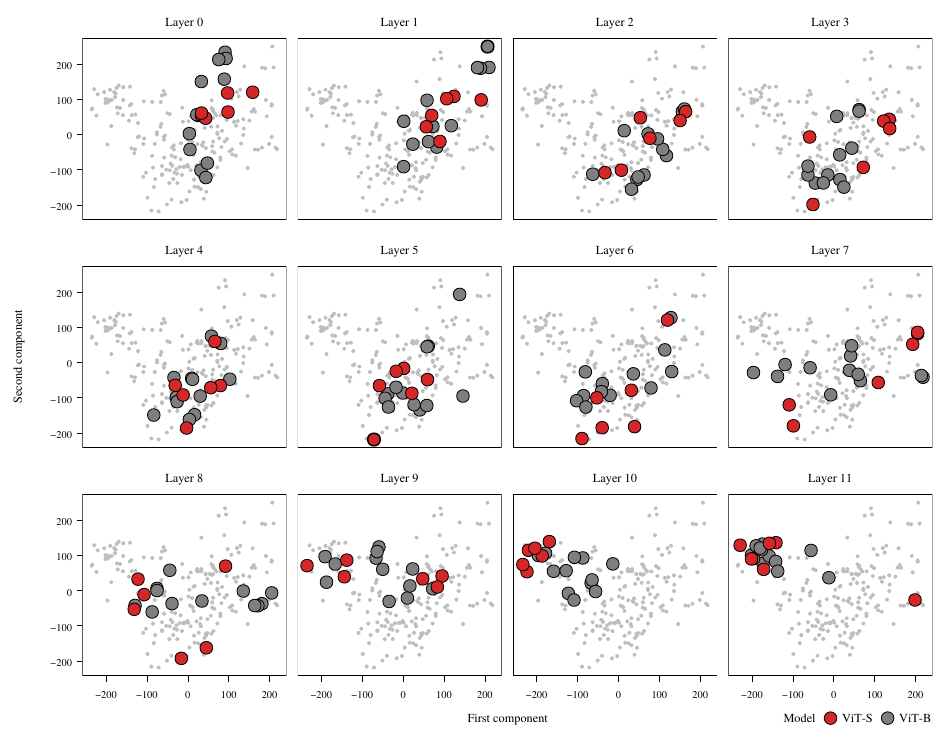}
  \caption{Extended results of the Principal Component Analysis performed on a common set of heads' activations from both ViT-S and ViT-B on the validation set of ImageNet-1K. In each subplot, the heads of that layer are marked in red (ViT-S) and gray (ViT-B), and the smaller gray dots are the heads of all other layers. Dependencies in the coordinates of the two main components between most of the subsequent layer heads suggest a visible relationship.}
  \label{fig:PCA_both_models_individual_layers}
\end{figure}

Figure~\ref{fig:PCA_both_models_individual_layers} shows the results of the Principal Component Analysis based on the activations of the two considered ViT variants, with the heads of consecutive layers marked separately, clearly demonstrating dependencies between the positions of subsequent layer heads. For instance, the heads of layers 0 and 1 have high positive values of both components, while the heads of layers 9--11 are mainly located in the upper left corner of the images.

\clearpage
\section{Router heads}\label{app:sink}\label{app:eigvec}
This section extends Figure~\ref{fig:sink_heads}, i.e. presents the \textit{router} rule on various heads and images (\cref{app:router_rule}), the masking of outlying patches for all \textit{router} heads (\cref{app:router_masking}), and the same analysis for DINOv2 with registers (\cref{app:registers}).

\subsection{The router rule}\label{app:router_rule}
Each of the three Laplace map channels (eigenvectors 2, 3 and 4, starting from the smallest) is scaled to $[0,1]$ over the 256 patches, and the mode $S$ is the center of the highest-frequency bin out of 100 bins. The map passes the condition when at least 96\% of the patches of every channel lie within the interval $[S-0.1, S+0.1]$. A head counts as a \textit{router} when its map passes on at least 75\% of the 50{,}000 ImageNet-1K validation images.

Figure~\ref{fig:eigvec_channels_app_more_a}--\ref{fig:eigvec_channels_app_more_b} show the \emph{router} selection rule for selected images and ViT-B heads. Unlike \emph{router} heads, the \textit{convolutional} heads and heads responsible for more complex shapes spread patch values. In \textit{router} heads, nearly all patches of the three eigenvectors fall inside the window around the mode (96--100\%). L8H11 in the first image shows that a simple rule considering one channel above the threshold is not enough to detect a \emph{router} head.

\begin{figure*}[!ht]
  \centering
  \includegraphics{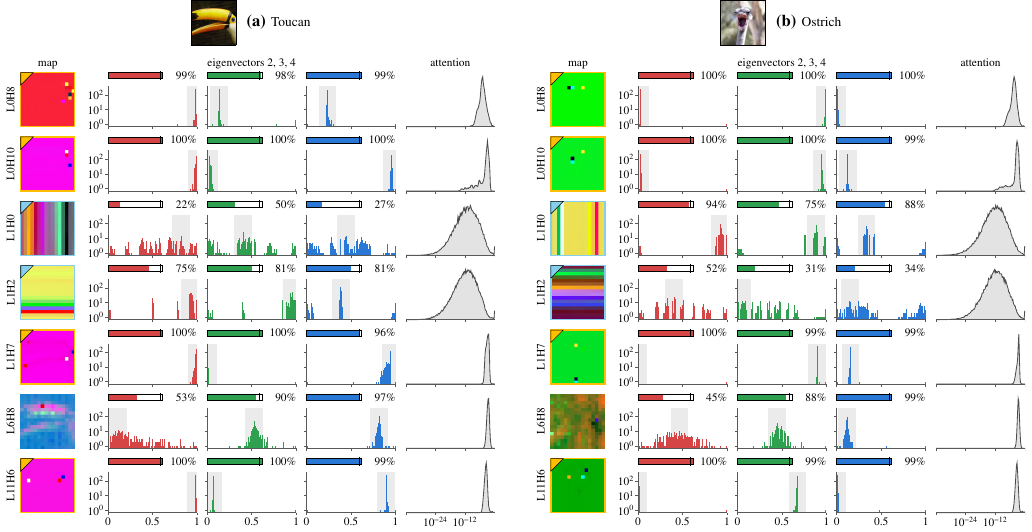}
  \caption{For each head, we present its Laplace map, the histograms of the 2nd--4th smallest eigenvectors over the 256 patches, with the $\pm0.1$ window around the mode (gray) and, above each, the share of patches inside the window. We compare it with the density of the attention weights. A map passes the rule when all three channels reach at least 96\%. Over each block, we additionally show the input image. Maps framed and marked in the corner belong to \textit{router} (yellow) and \textit{convolutional} (light blue) heads.}
  \label{fig:eigvec_channels_app_more_a}
\end{figure*}

\begin{figure*}[p]
  \centering
  \includegraphics{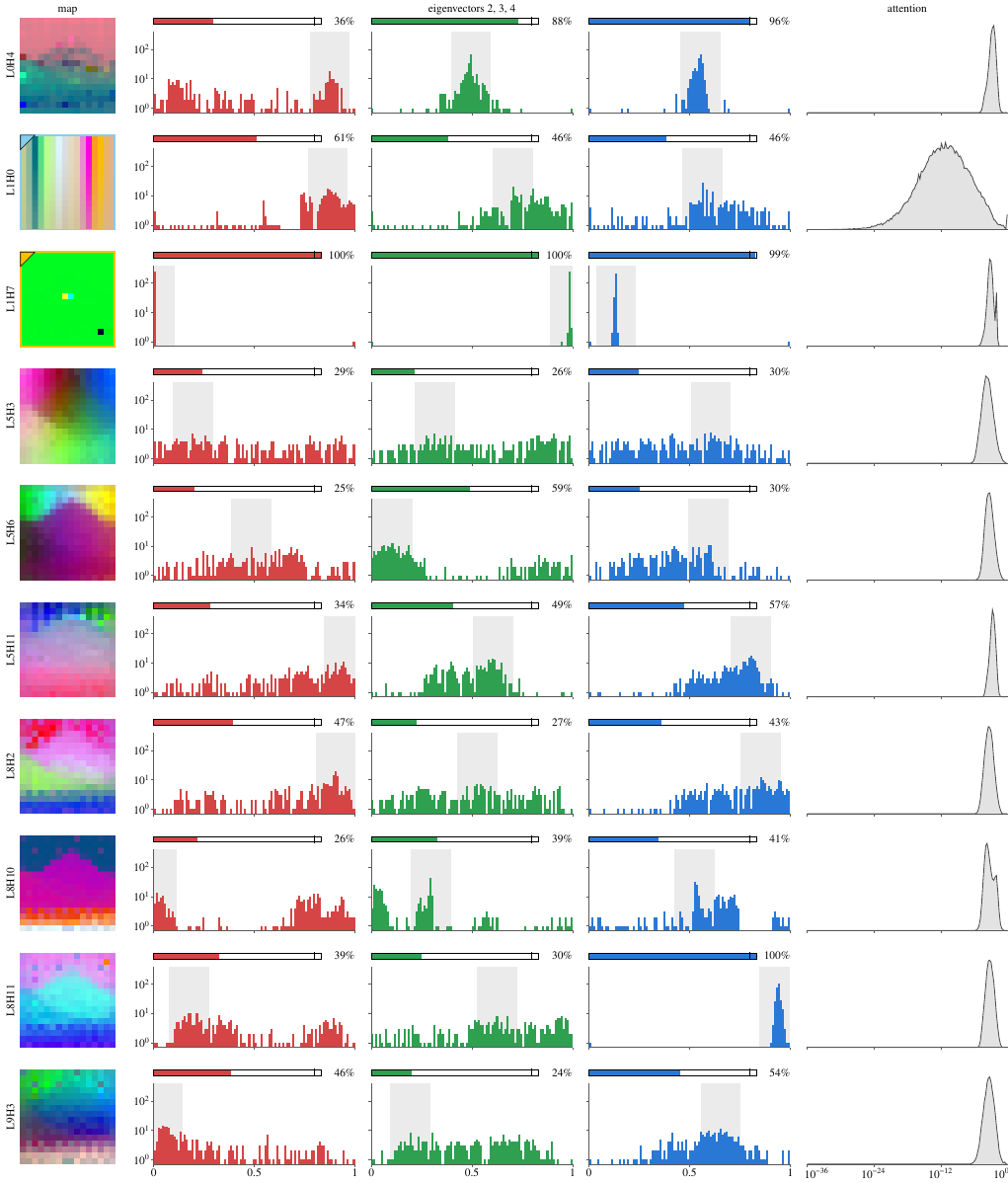}
  \caption{Continuation of Figure~\ref{fig:eigvec_channels_app_more_a} for
  ten further ViT-B heads on the image of Figure~\ref{fig:eigvec_channels}. The density of the head's $256\times256$ patch-to-patch attention weights is presented on a log scale.}
  \label{fig:eigvec_channels_app_volcano}
\end{figure*}

\begin{figure*}[p]
  \centering
  \includegraphics{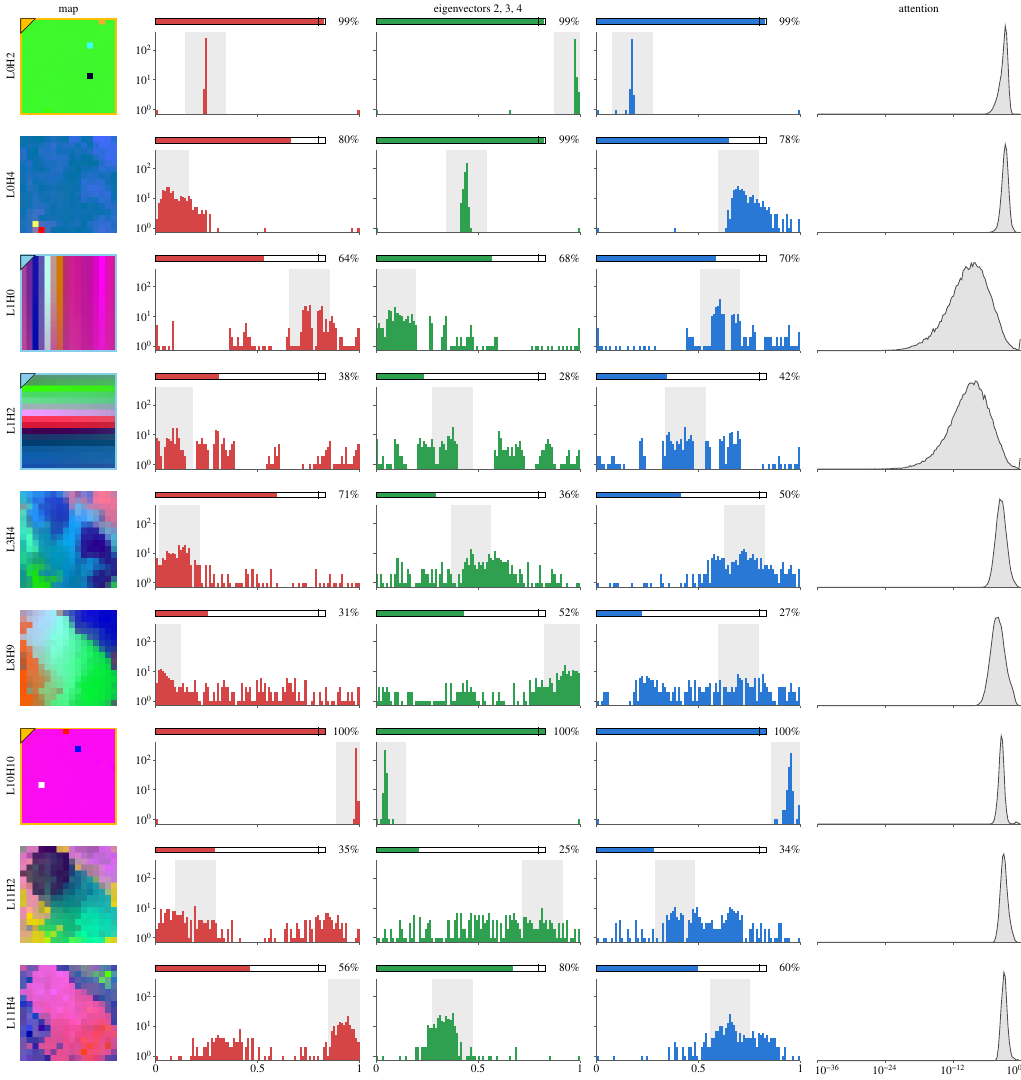}
  \caption{Continuation of Figures~\ref{fig:eigvec_channels_app_more_a}--\ref{fig:eigvec_channels_app_volcano} for
  nine ViT-B heads on a second image.}
  \label{fig:eigvec_channels_app_spaniel}
\end{figure*}

Figures~\ref{fig:eigvec_channels_app_more_vit_s_a} and~\ref{fig:eigvec_channels_app_more_vit_s_b} show ViT-S on eight images: L0H4 and L1H3 fulfill the condition on every image, L0H1 fails on the toucan (95\%) and on the traffic light (45\%), and L9H4 fails on the volcano (94\%). Also, between 7 and 11 of the 11 \textit{router} heads pass a condition on a single image.

\begin{figure*}[p]
  \centering
  \includegraphics{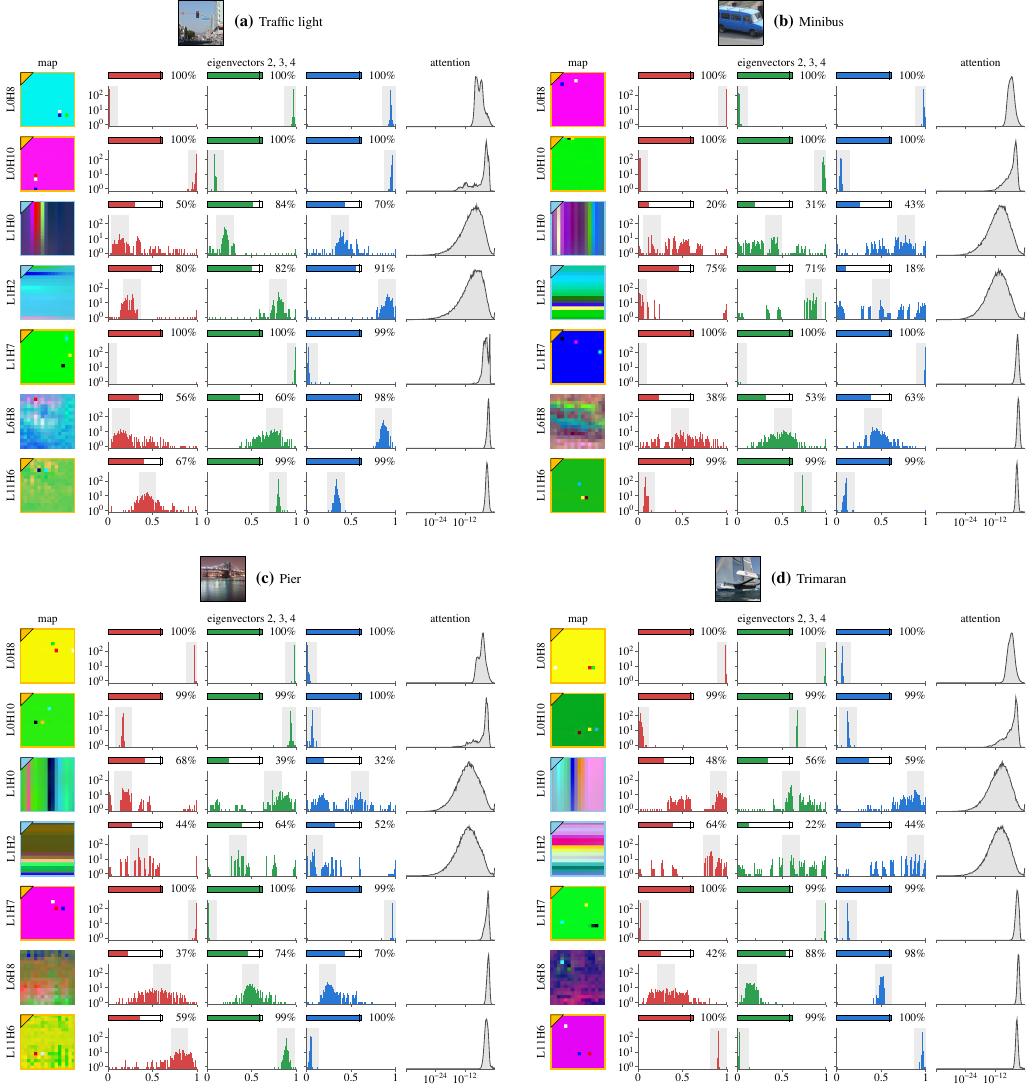}
  \caption{The \textit{router} rule applied to four more images, for the heads presented in Figure~\ref{fig:eigvec_channels_app_more_a}. The \textit{router} head L11H6 fails the rule on the traffic light (\textbf{a}, 67\%) and on the pier (\textbf{c}, 59\%) and passes on the other four of the six images. We set that a head counts as a \textit{router} when it meets the condition on at least 75\% of images.}
  \label{fig:eigvec_channels_app_more_b}
\end{figure*}

\begin{figure*}[p]
  \centering
  \includegraphics{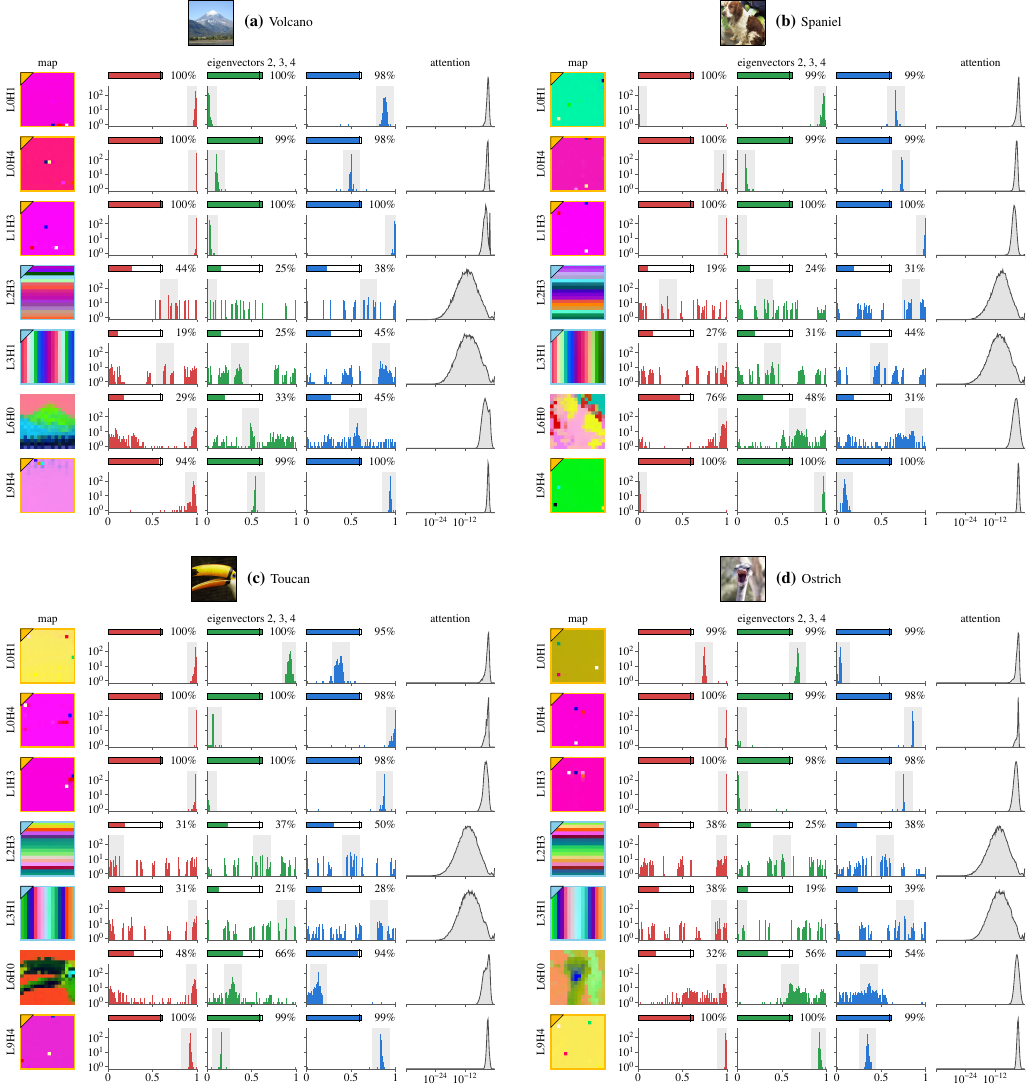}
  \caption{Continuation of Figure~\ref{fig:eigvec_channels_app_more_b}, but for ViT-S heads. Two \textit{router} heads meet the condition on every image (L0H4, L1H3), while two fail on some (L0H1 on the toucan, L9H4 on the volcano). We also present the \textit{convolutional} heads (horizontal L2H3 and vertical L3H1) and an ordinary head (L6H0).}
  \label{fig:eigvec_channels_app_more_vit_s_a}
\end{figure*}
\begin{figure*}[p]
  \centering
  \includegraphics{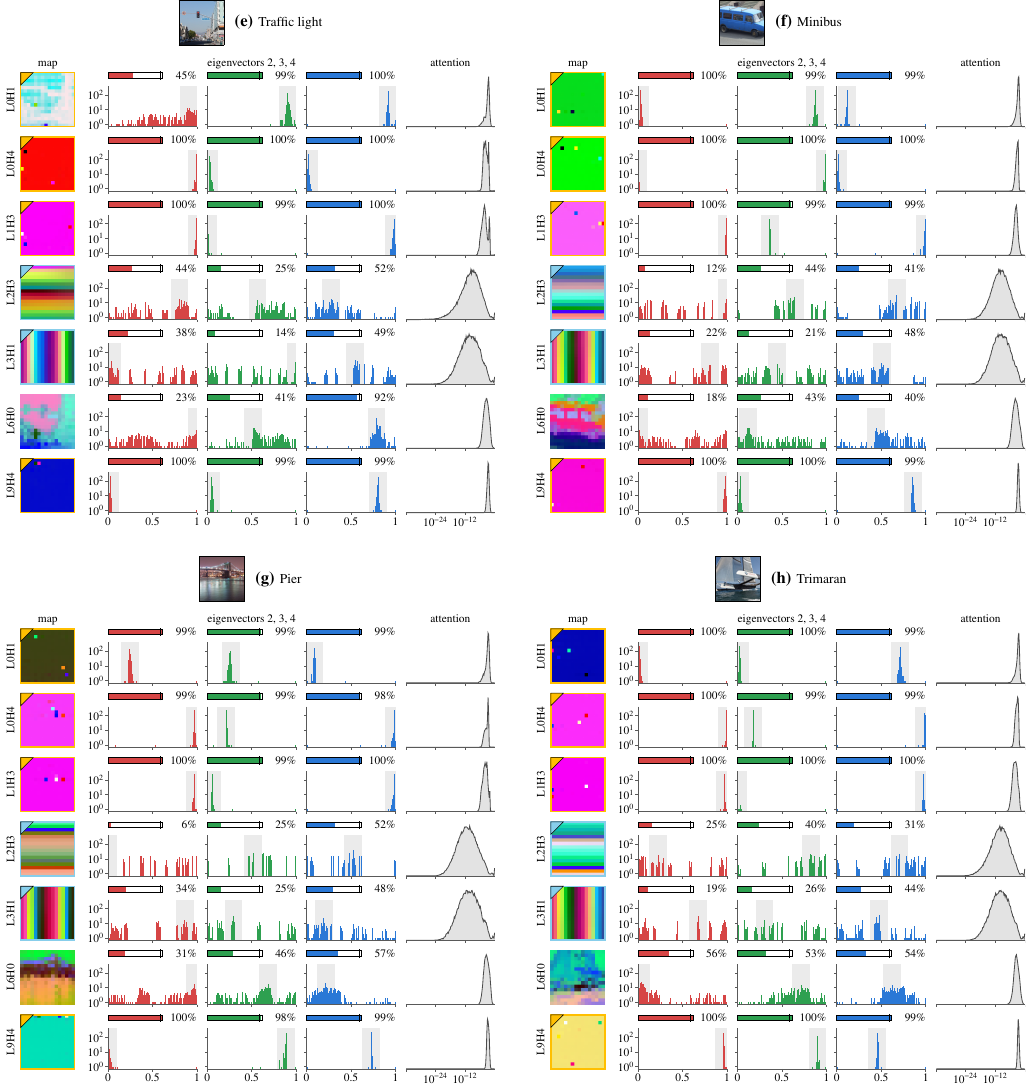}
  \caption{Continuation of Figures~\ref{fig:eigvec_channels_app_more_b}--\ref{fig:eigvec_channels_app_more_vit_s_a}, ViT-S heads.}
  \label{fig:eigvec_channels_app_more_vit_s_b}
\end{figure*}

\FloatBarrier

\subsection{Masking outlying patches}\label{app:router_masking}
\begin{figure*}[!h]
  \centering
  \includegraphics{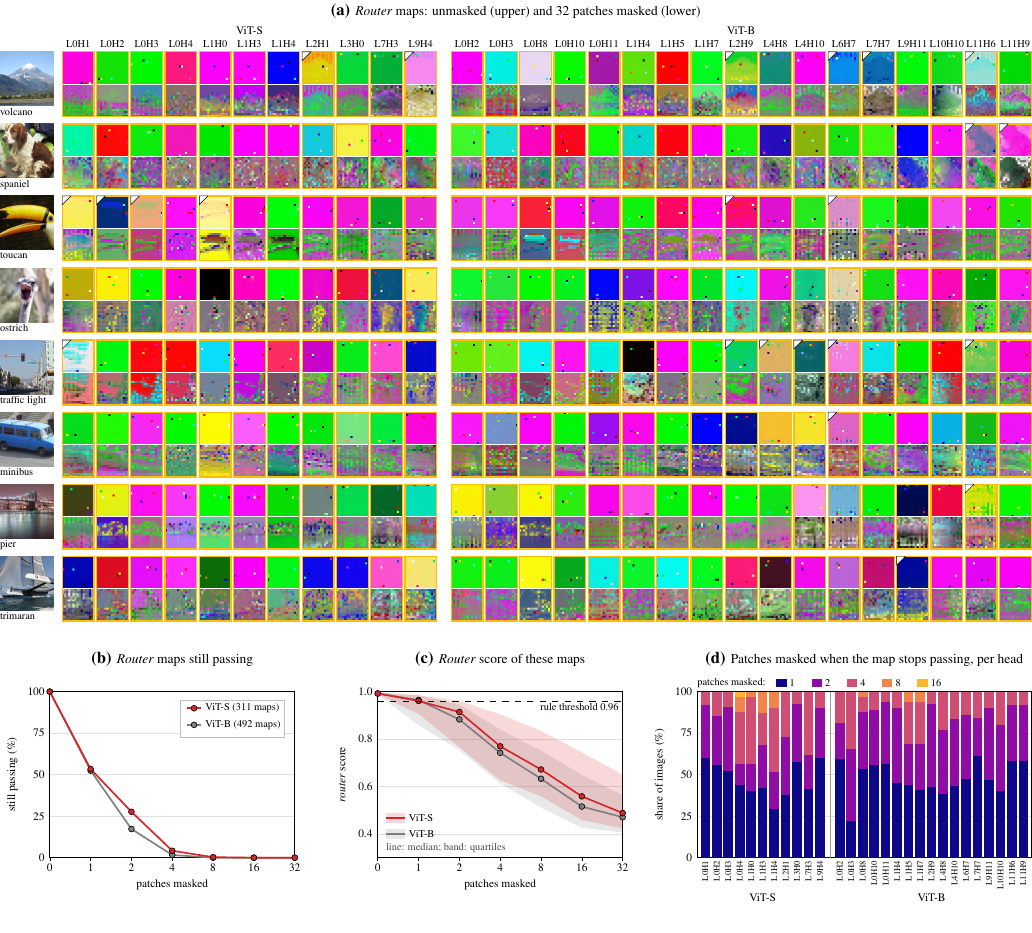}
  \caption{Masking the outlying patches of the \emph{router} heads, as presented in Figure~\ref{fig:sink_heads}c). \textbf{(a)} Every \emph{router} head of both models, before and after masking its 32 most outlying patches. A white corner marks a head that does not fulfill the rule on that image. Over 32 images, for the \emph{router} maps that pass the rule when heads were unmasked, we also show \textbf{(b)} the percentage of heads that still meet the condition, having $n$ masked patches masked. \textbf{(c)} presents the \emph{router} score over the three channels against the threshold of the rule. Finally, \textbf{(d)} depicts the number of images on which a \emph{router} head's map stops fulfilling the condition after masking 1, 2, 4, 8 or 16 patches.}
  \label{fig:app_masking_images}
\end{figure*}

Figure~\ref{fig:app_masking_images} introduces the masking of Figure~\ref{fig:sink_heads}c) for every \textit{router} head of ViT-S and ViT-B, while Figures~\ref{fig:app_masking_ostrich_vit_s} (ViT-S) and~\ref{fig:app_masking_ostrich_vit_b} (ViT-B) present their Laplace maps where 1 to 32 of the outlying patches are masked. Masking a single patch is enough for about half of the \textit{router} heads to stop passing the rule, while eight masked patches led to almost no passes (0.3\% of ViT-S maps and none of ViT-B, over the 32 images). In general, the \textit{router} head's map concentrates on a handful of outlying patches. Here, we reveal that once such patches are masked, the object appears. It may be caused, e.g., by residual connections from other heads, and this observation needs further verification. The median \textit{router} score falls to the threshold with one patch masked, and every map stops passing by 16 patches.

\begin{figure*}[p]
  \centering
  \includegraphics{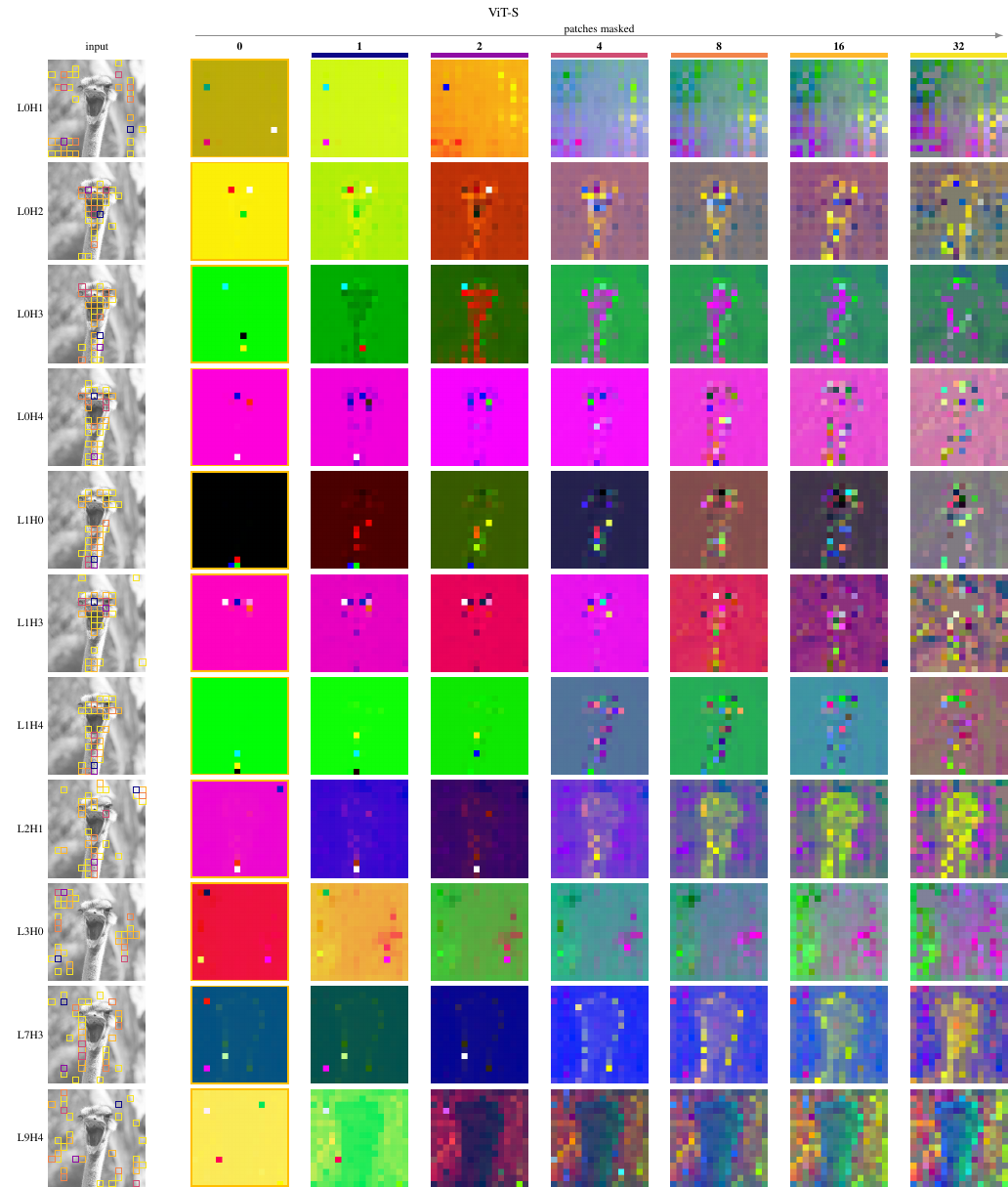}
  \caption{Masking the outlying patches of every \textit{router} head of ViT-S on the ostrich of Figure~\ref{fig:sink_heads}c. The outer left column presents the input, with the 32 most outlying patches of the head marked in the color describing the level at which a given patch is masked. The next columns present the head's Laplace map with 0 to 32 of these patches masked. Each masked patch is set to the mean of the other patches in each eigenvector, and the eigenvectors are not recomputed.}
  \label{fig:app_masking_ostrich}\label{fig:app_masking_ostrich_vit_s}\label{fig:app_sink_masking_vit_s}
\end{figure*}
\begin{figure*}[p]
  \centering
  \includegraphics{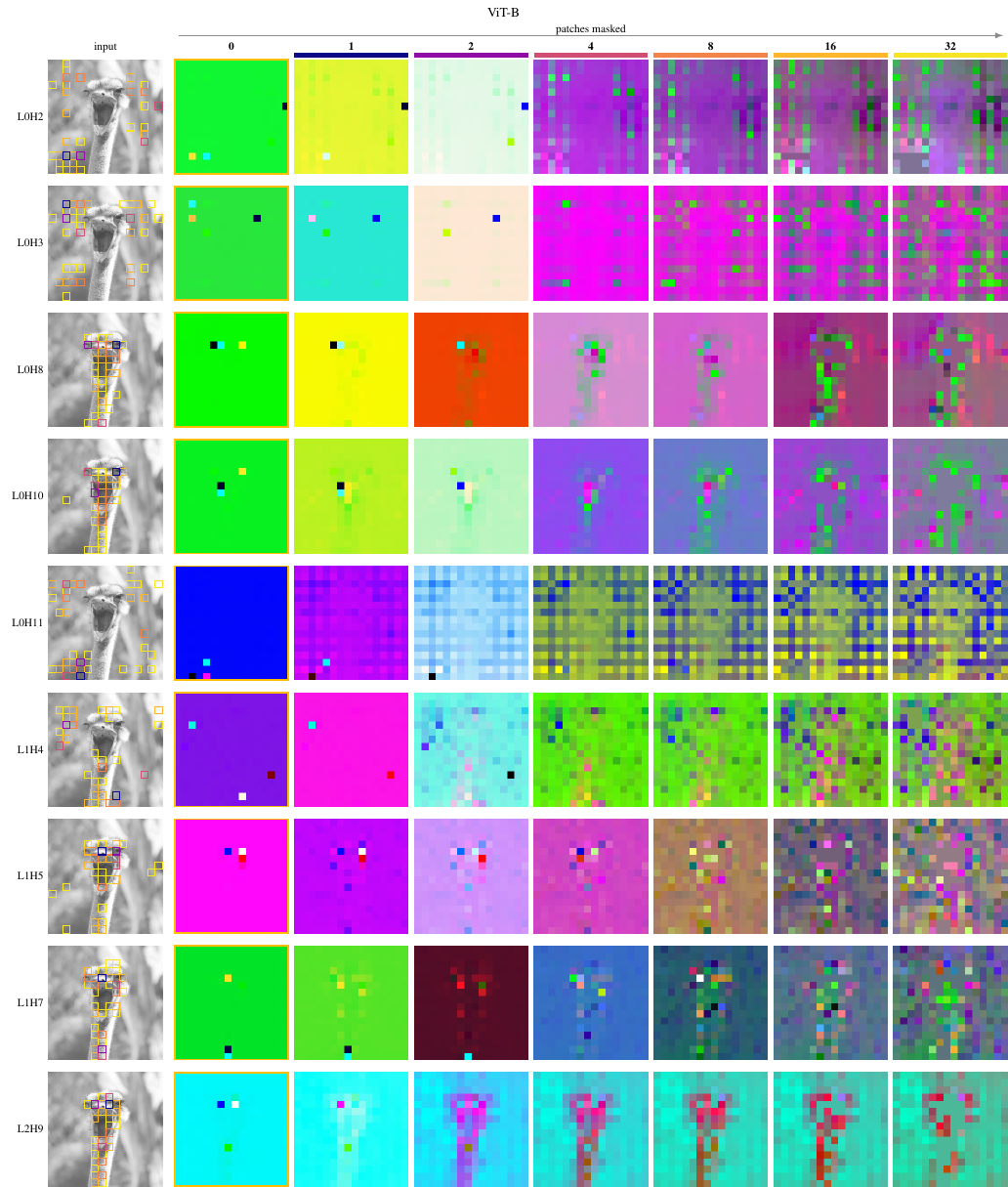}
  \caption{Continuation of Figure~\ref{fig:app_masking_ostrich_vit_s}, \emph{router} heads of ViT-B in layers 0--2.}
  \label{fig:app_masking_ostrich_vit_b}\label{fig:app_sink_masking_vit_b}
\end{figure*}
\begin{figure*}[p]
  \centering
\includegraphics{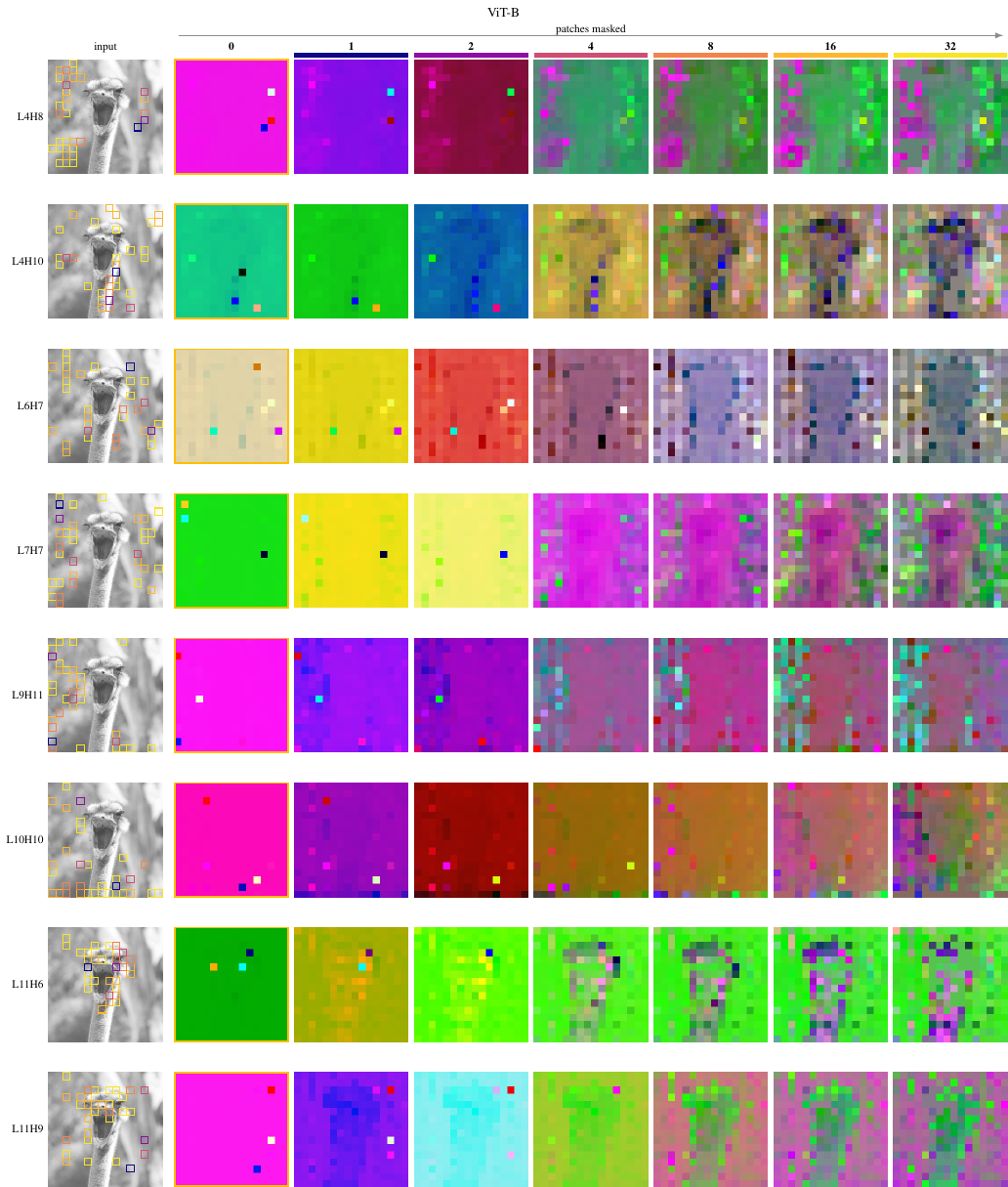}
  \caption[]{Continuation of Figures~\ref{fig:app_masking_images}--\ref{fig:app_masking_ostrich_vit_b}, \emph{router} heads of ViT-B in layers 4--11.}
\end{figure*}

\FloatBarrier

\subsection{DINOv2 with registers}\label{app:registers}
\begin{figure}[!h]
  \centering
  \begin{subfigure}[b]{\linewidth}\phantomsubcaption\label{fig:app_registers_share}\end{subfigure}\begin{subfigure}[b]{\linewidth}\phantomsubcaption\label{fig:app_registers_share_plain}\end{subfigure}
  \includegraphics[width=\linewidth]{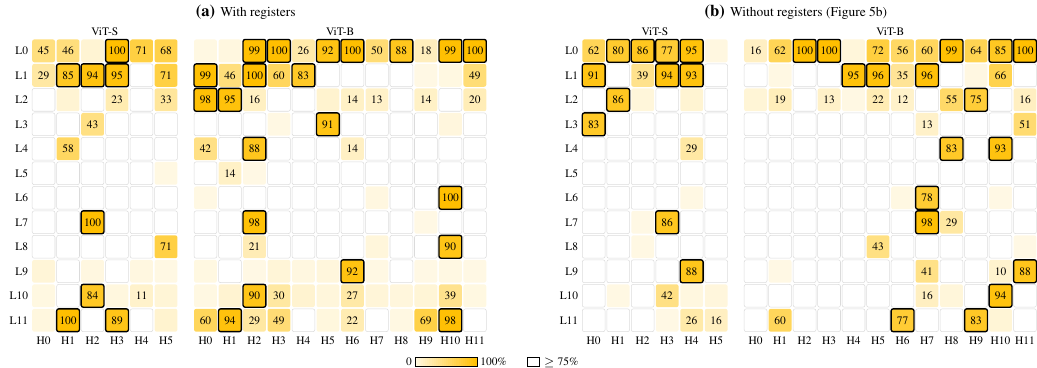}
  \caption{The percentage of the 50{,}000 ImageNet-1K validation images on which each head passes the \textit{router} rule, (a) in DINOv2 with registers (CLS and the four registers dropped after the softmax) and (b) without registers, as in Figure~\ref{fig:sink_heads}b. Heads that meet the condition in at least 75\%, i.e.\ \emph{routers}, are framed: 8 in ViT-S and 21 in ViT-B with registers, 11 and 17 without.}
  \label{fig:app_registers}
\end{figure}

In DINOv2 with registers, the rule still isolates \emph{router} heads (Figure~\ref{fig:app_registers_share}): 8 in ViT-S and 21 in ViT-B, against 11 and 17 without registers (Figure~\ref{fig:app_registers_share_plain}, the data of Figure~\ref{fig:sink_heads}b). Only selected heads are \emph{routers} with and without registers: L0H3 and L1H3 in ViT-S, and six heads of the first two layers in ViT-B (L0H2, L0H3, L0H8, L0H10, L0H11, L1H4). With registers, \textit{router} heads also appear in the last layers (ViT-S L10H2, L11H1, L11H3; ViT-B L10H2, L11H1, L11H10).

\begin{figure*}[!h]
  \centering
  \includegraphics{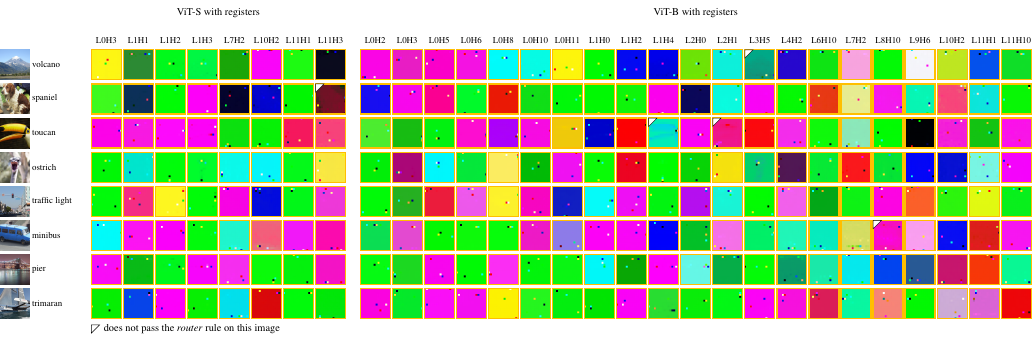}
  \caption{The \textit{router} heads of DINOv2 with registers (8 in ViT-S, 21 in ViT-B) on the eight selected images. A white corner marks a map that does not fulfill the rule on that image (1 of 64 for ViT-S, 4 of 168 for ViT-B). Figure~\ref{fig:app_registers_maps_reg} shows every head of these models on the ostrich.}
  \label{fig:app_registers_more}
\end{figure*}

\noindent Figure~\ref{fig:app_registers_more} shows the 29 \textit{router} heads of the register models on eight images: 1 of the 64 ViT-S maps and 4 of the 168 ViT-B heads do not fulfill the rule on this subset of images.

\begin{figure*}[p]
  \centering
  \begin{subfigure}[b]{\linewidth}\phantomsubcaption\label{fig:app_registers_maps_reg}\end{subfigure}\begin{subfigure}[b]{\linewidth}\phantomsubcaption\label{fig:app_registers_maps_plain}\end{subfigure}
  \includegraphics[width=\linewidth]{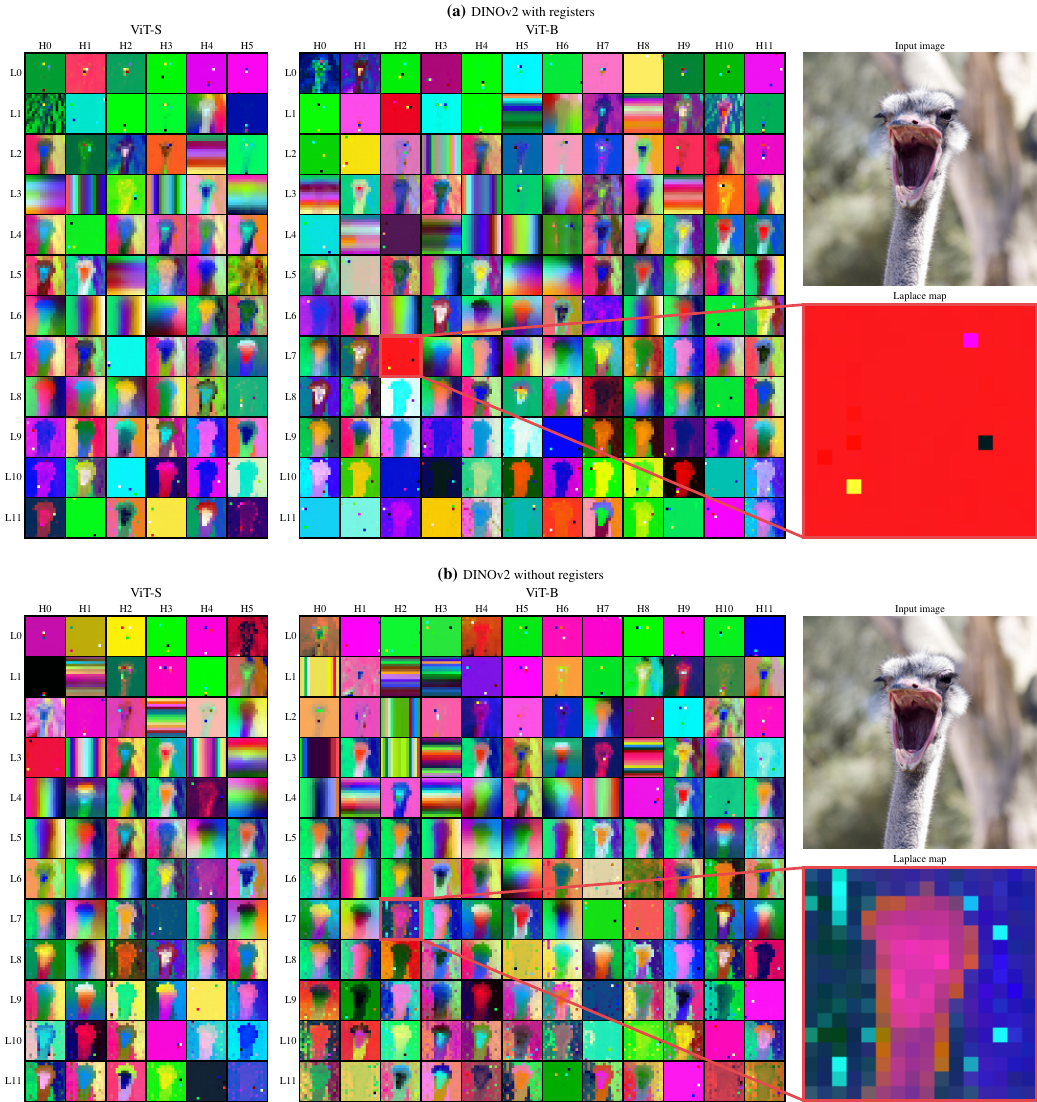}
  \caption{Laplace maps of all heads of ViT-S and ViT-B on the image from Figure~\ref{fig:sink_heads}c, \textbf{(a)} with registers and \textbf{(b)} without, as in Figure~\ref{laplace_maps_full_3}. The outer right column presents the input image and the enlarged L7H2 head for ViT-B (red), passing the rule for 97.5\% of the images. This head is semantic without registers (0.01\% of images).}
  \label{fig:app_registers_maps}
\end{figure*}

\noindent Figure~\ref{fig:app_registers_maps} shows every head in the image of Figure~\ref{fig:sink_heads}c for models with registers \textbf{(a)} and without \textbf{(b)}, where all 29 and all 28 \textit{router} heads pass the rule. L7H2, enlarged there, is a \textit{router} head with registers in both model size variants (99.8\% of the images in ViT-S, 97.5\% in ViT-B) and a semantic head without registers.

\FloatBarrier   

\clearpage

\section{Convolutional heads}\label{app:conv}
\begin{figure}[!ht]
  \centering
  \includegraphics{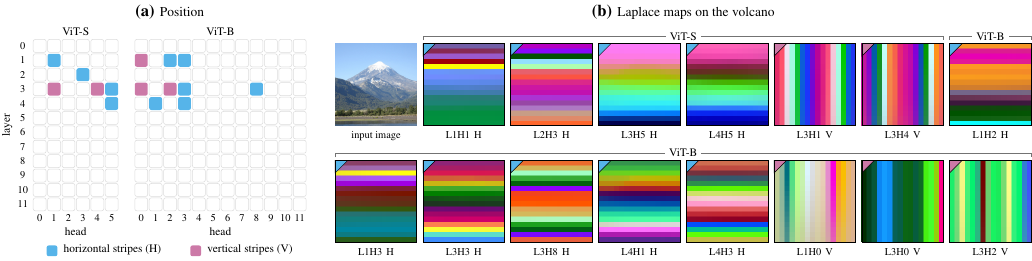}
  \caption{\textit{Convolutional} heads of DINOv2, i.e. heads whose Laplace map consists of horizontal (H) or vertical (V) stripes, so that all patches of a row (or of a column) receive the same value. \textbf{(a)} Location of these heads in ViT-S and ViT-B (4H $\backslash$ 2V, and 6H $\backslash$ 3V heads, respectively, all in layers L1--L4). \textbf{(b)} The Laplace map of each of them on the image of Figure~\ref{laplace_maps_full_1} (top), with the input image first, then the heads of ViT-S and of ViT-B.}
  \label{fig:app_conv_heads}
\end{figure}

Besides the router heads, we can identify a group of \emph{convolutional} heads, for which a Laplace map consists of horizontal or vertical stripes, so that all patches of a row (or a column) receive the same or comparable value. Such a head treats the image like a one-dimensional filter along a single axis. Figure~\ref{fig:app_conv_heads} presents these heads, i.e. 4 horizontal and 2 vertical in ViT-S, 6 horizontal and 3 vertical in ViT-B, all located in layers L1 to L4.

\subsection{Sobel filtering}
Our approach relies on finding horizontal/vertical edges with Sobel filtering. After preprocessing steps for the head $h$ in the layer $i$, performed in the same way as for semantic clustering, leading to $\mathbf{\hat{A}}_i^h$, the attention should be reshaped into a grid $16 \times 16$, according to the number of patches in DINOv2. For simplicity, we omit the layer and head indices. Then, two kernels should be convolved with the attention map $\mathbf{A}$:

\begin{equation}
    \mathbf{G_x} = \begin{bmatrix}
        -1 & 0 & 1 \\
        -2 & 0 & 2 \\
        -1 & 0 & 1 \\
    \end{bmatrix} * \mathbf{A}.
\end{equation}

\begin{equation}
    \mathbf{G_y} = \begin{bmatrix}
        -1 & -2 & -1 \\
        0 & 0 & 0 \\
        1 & 2 & 1 \\
    \end{bmatrix} * \mathbf{A}.
\end{equation}

\noindent With $\mathbf{G_x}$ and $\mathbf{G_y}$ we can calculate energy for this head in horizontal $(Hor)$ and vertical $(Ver)$ directions:

\begin{equation}
Hor = \sum\limits_{i=1}^{N} \sum\limits_{p \in \mathbf{G_x}} |\mathbf{G_x}(i, p)|,
\end{equation}

\begin{equation}
Ver = \sum\limits_{i=1}^{N} \sum\limits_{p \in \mathbf{G_y}} |\mathbf{G_y}(i, p)|.
\end{equation}

\noindent Finally, the ratio $R$ will assess whether the head is focused primarily on horizontal edges $(R \approx 1)$, vertical edges $(R \approx -1)$, or more sophisticated shapes $(R \approx 0)$. Finally, to identify a convolutional head, we calculate the total energy in both directions:
\begin{equation}
    R = \dfrac{Hor - Ver}{Hor + Ver}.
\end{equation}

\newpage
\subsection{Identification results}

\begin{figure*}[!ht]
  \centering
  \includegraphics{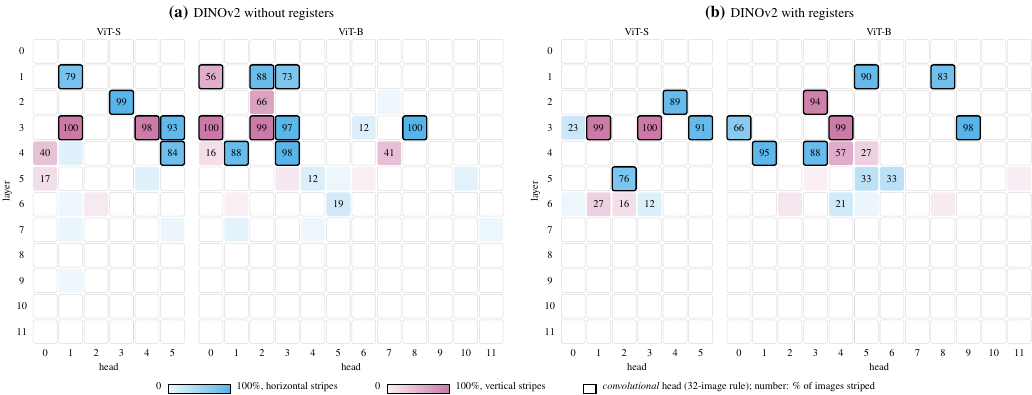}
  \caption{\textit{Convolutional} heads on all 50{,}000 ImageNet-1K validation images for DINOv2 \textbf{(a)} without and \textbf{(b)} with registers. The numbers represent the percentage (\%) of images on which a head's Laplace map is striped. Blue color represents horizontal stripes, while purple corresponds to vertical ones.}
  \label{fig:app_conv_share}
\end{figure*}

Figure~\ref{fig:app_conv_share} identifies the \emph{convolutional} head type on all 50{,}000 validation images, both for the models with and without registers. Every \emph{convolutional} head is striped for most images (from 56\% for L1H0 of ViT-B to 100\%). With registers, 3 \emph{convolutional} heads are horizontal in ViT-B, and 2 are vertical in ViT-S, as well as 6 and 2 in ViT-B, respectively.

\section{What remains after pruning}\label{app:remains}
\begin{figure}[!ht]
  \centering
  \includegraphics{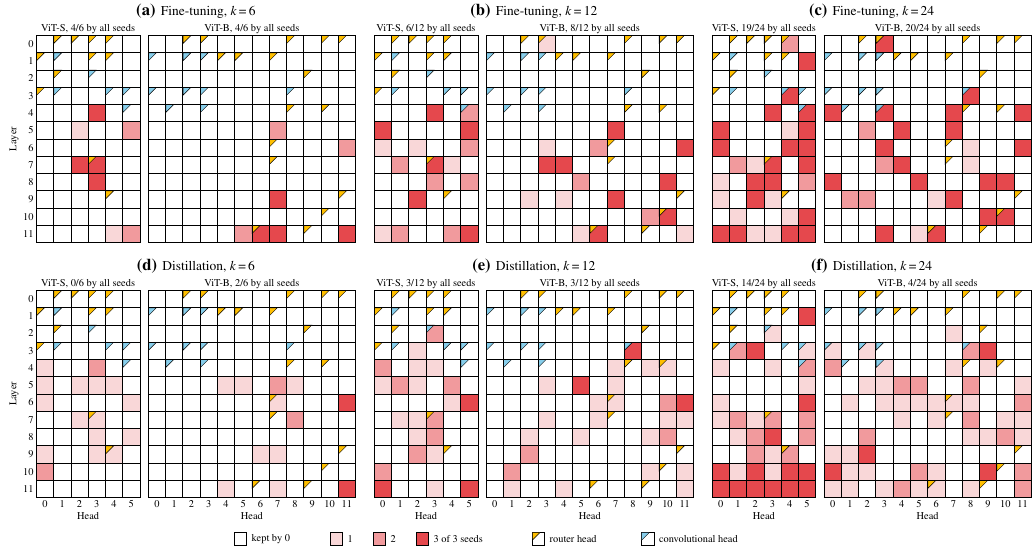}
  \caption{Head selection for all settings across runs (Figure~\ref{fig:pruned_models}a extended) for the best models (over 3 seeds) after fine-tuning (top) and distillation (bottom), for $k = 6, 12$ and $24$. Corner marks denote \textit{router} heads (amber) and \textit{convolutional} heads (light blue).}
  \label{fig:app_remains_settings}
\end{figure}

\begin{figure*}[p]
  \centering
  \includegraphics{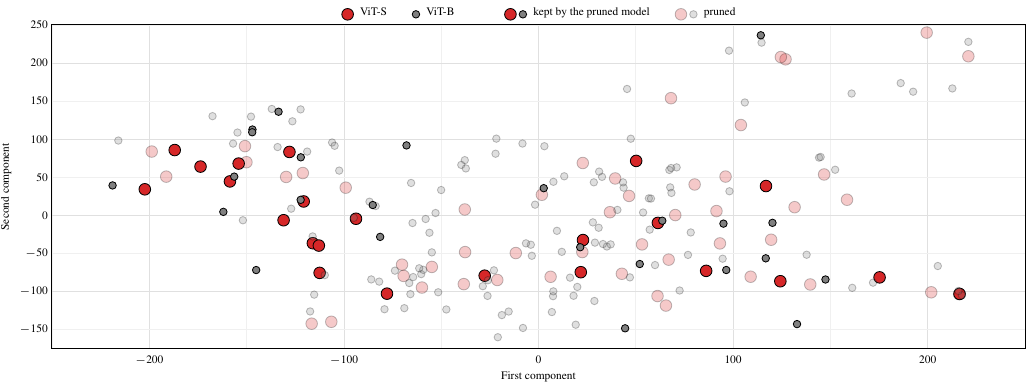}
  \caption{Retained heads in the joint PCA of all heads of ViT-S and ViT-B: the points of Figure~\ref{fig:pruned_models}b, drawn over the full width. Opaque markers correspond to the heads marked as retained there (23 of ViT-S, 24 of ViT-B), while faded markers denote the others.}
  \label{fig:app_remains_pca}
\end{figure*}

In this section, we analyze the distribution of heads that were selected by \our{} in various experimental settings and seeds.
We present exemplary model runs trained according to the best hyperparameter sets, with three seeds per setting. 
After fine-tuning, most retained heads are shared by all three seeds (19 of 24 in ViT-S and 20 of 24 in ViT-B at $k=24$), while after distillation the seeds largely disagree (14 of 24 and 4 of 24, respectively). For the extreme case, i.e. $k=6$, there are no common heads for three seeds after distillation in ViT-S, while 2 out of 6 heads were common in ViT-B, as presented in Figure~\ref{fig:app_remains_settings}. Figures~\ref{fig:app_importance_k6}--~\ref{fig:app_importance_k24} extend Figure~\ref{fig:pruned_models}c to both training modes and both model variants for $k=6$, $12$ and $24$, while Figures~\ref{fig:app_seed_drop}--\ref{fig:app_seed_drop_ft} present the accuracy lost when we remove a selected head for a distilled and a fine-tuned ViT-B across all seeds combined within a given pruning setting. Finally, Figure~\ref{fig:app_remains_pca} presents the retained heads in the PCA plot for all heads.

\begin{figure*}[!ht]
  \centering
\includegraphics{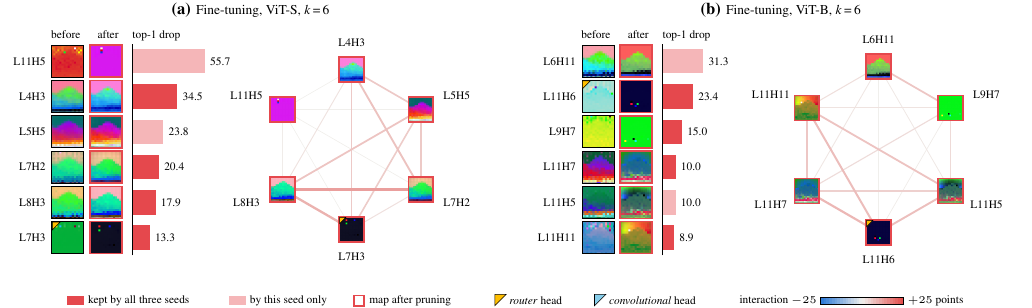}
  \caption{Head importance and interactions at $k = 6$ for fine-tuning, seed 42.}
  \label{fig:app_importance_k6}
\end{figure*}
\begin{figure*}[!ht]
  \centering
\includegraphics{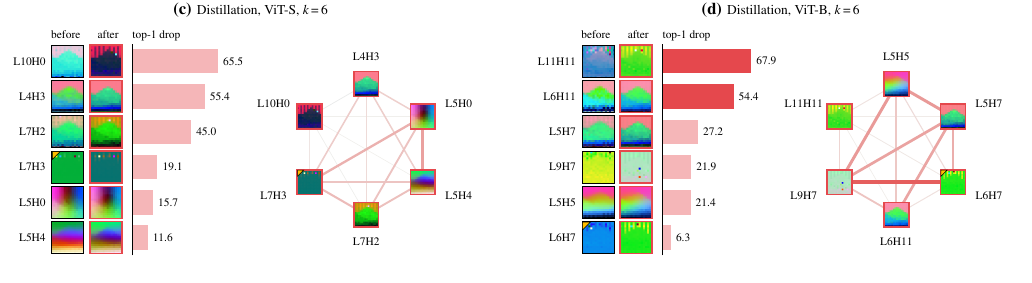}
  \caption[]{Continuation of Figure~\ref{fig:app_importance_k6}, $k=6$, distillation.}
\end{figure*}

\begin{figure}[!ht]
  \centering
  \includegraphics{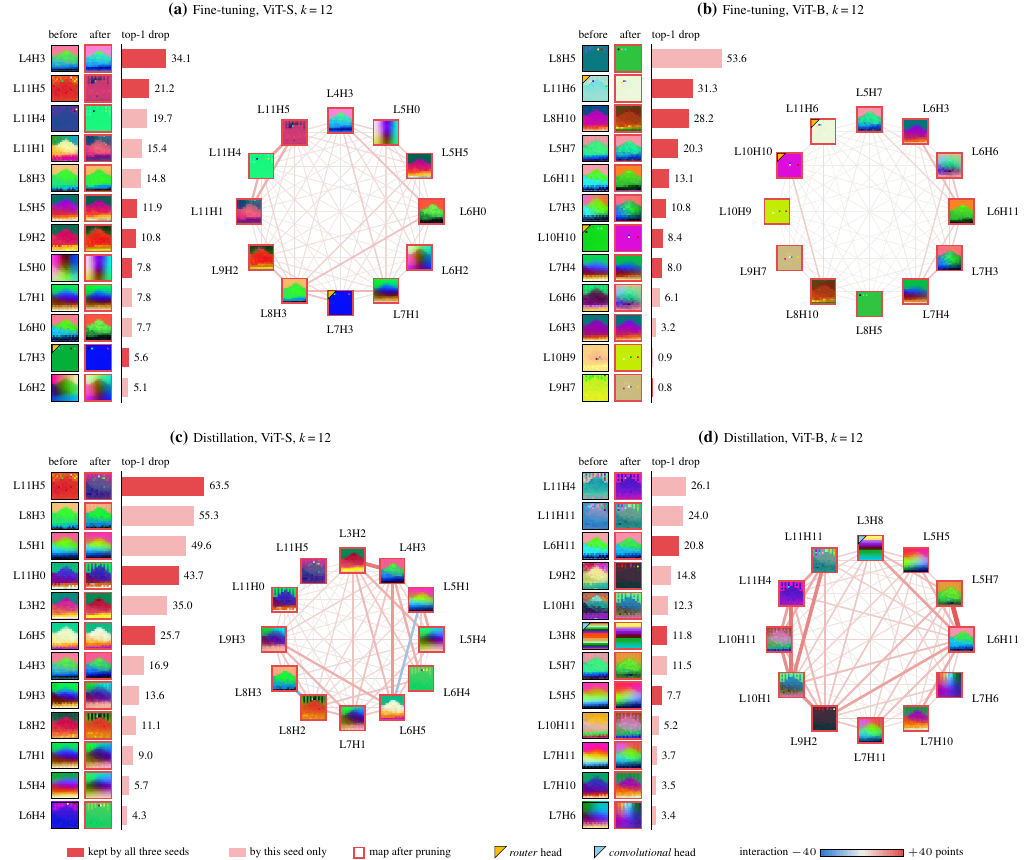}
  \caption{Head importance interaction plots for both training modes and model sizes for an exemplary hyperparameter set at $k = 12$ (Figure~\ref{fig:pruned_models}c extended with one model per setting, seed 42. all seeds share the same hyperparameters). The retained heads ranked by the top-1 drop (points) when one is removed, with the Laplace map before and after pruning. Dark bars mark heads retained by all three seeds. Right columns present the interaction of every pair of the retained heads. The blue color indicates that the pair costs less than its two heads alone, while the red color means more. All panels share the drop and the interaction scale.}
  \label{fig:app_importance}
\end{figure}

\begin{figure*}[p]
  \centering
  \includegraphics{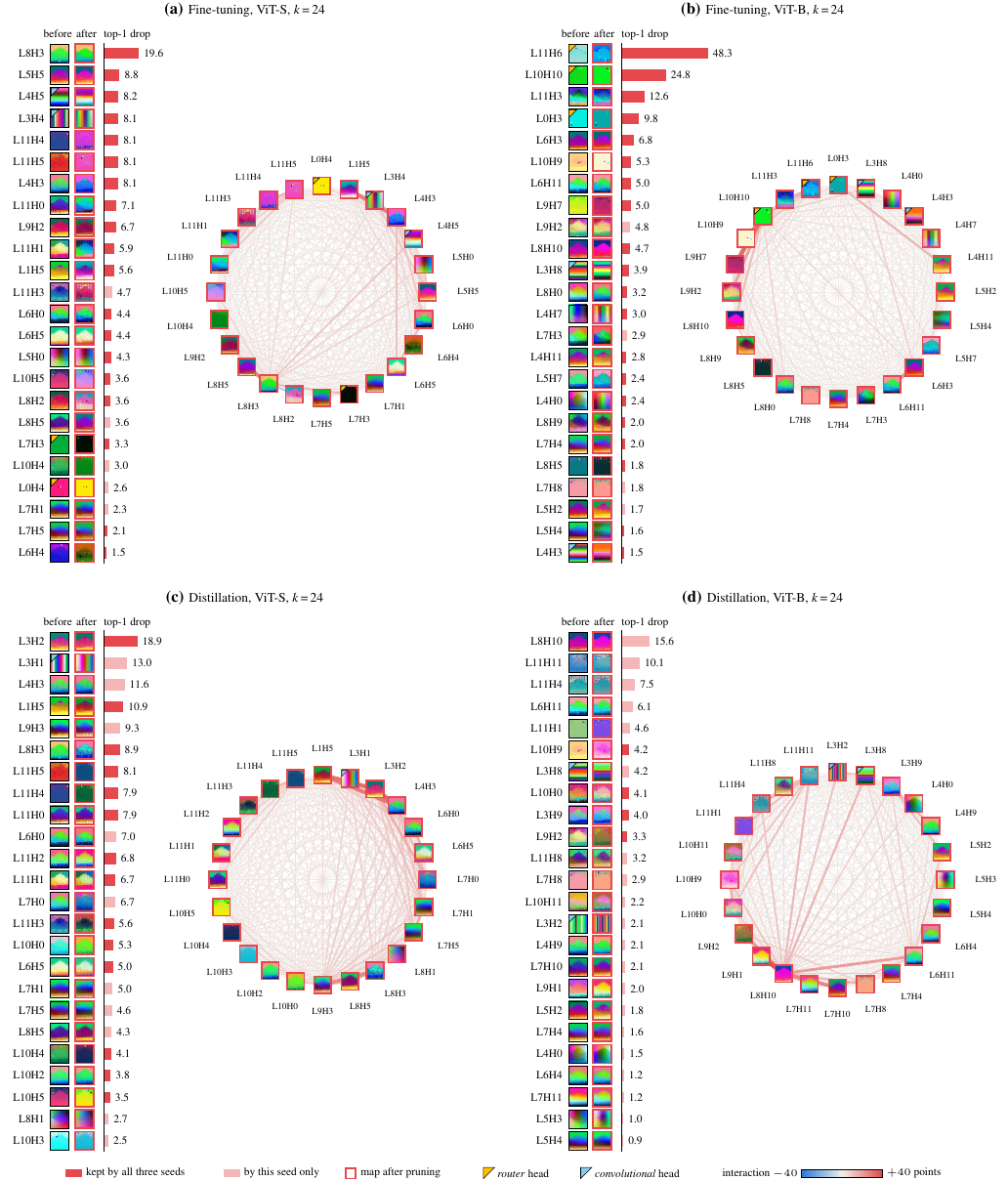}
  \caption{Continuation of Figure~\ref{fig:app_importance}, at $k = 24$: head importance and the interaction of all 276 pairs of the retained heads, for both training modes and both models. Scales are common for all four panels.}
  \label{fig:app_importance_k24}
\end{figure*}

\begin{figure*}[p]
  \centering
  \includegraphics{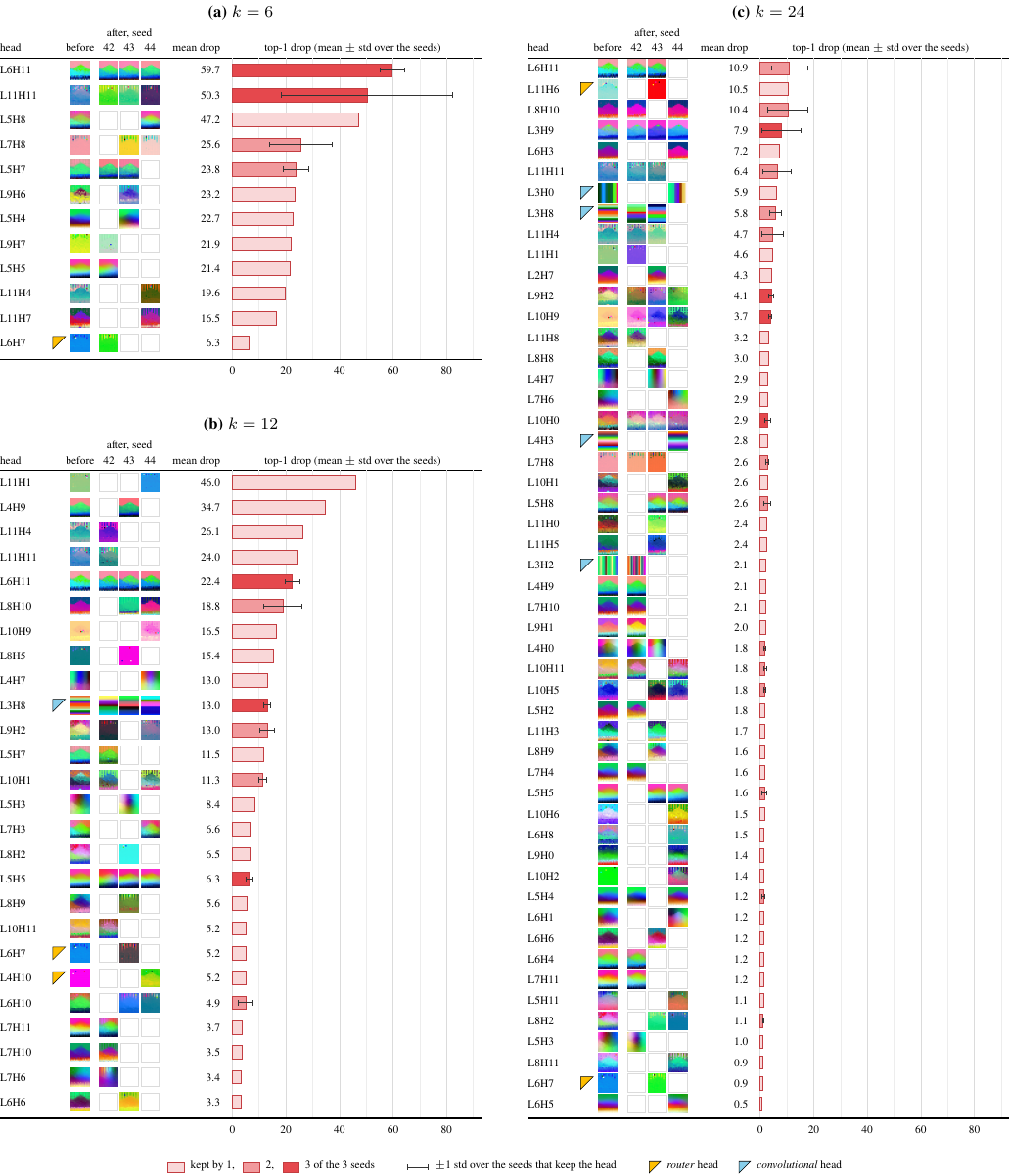}
  \caption{Results of the top-1 drop when one retained head is removed (distillation, ViT-B) over three seeds at (a) $k=6$, (b) $k=12$ and (c) $k=24$. The list presents every head that was retained by at least one seed, and is ranked by its mean drop. For each head, its Laplace map before pruning and after pruning in each seed's model is presented, while an empty box means that the head was not selected by that seed. Furthermore, we present the mean drop over the seeds for which it was selected as one of the best-performing heads. We also describe the standard deviation over those seeds or the lack of whiskers for a head kept by a single seed. As previously, corner marks denote \textit{router} heads (amber) and \textit{convolutional} heads (light blue).}
  \label{fig:app_seed_drop}\label{fig:app_remains_distill}
\end{figure*}
\begin{figure*}[p]
  \centering
  \includegraphics{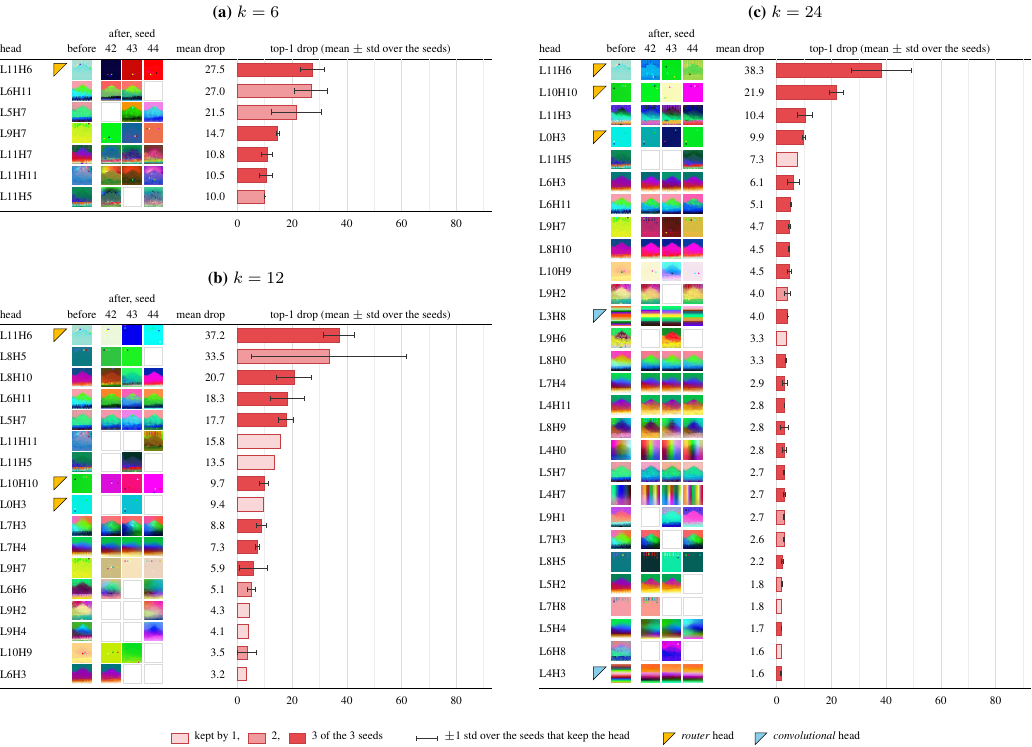}
  \caption{Continuation of Figure~\ref{fig:app_remains_distill} for fine-tuning on ViT-B at (a) $k=6$, (b) $k=12$ and (c) $k=24$.}
  \label{fig:app_seed_drop_ft}
\end{figure*}

\clearpage
\section{Extended results of pruning}\label{app:results}
Table~\ref{tbl:saper_extended} presents results of the \our{} comparison with the RAPTOR baseline across various experimental settings in a single place. FLOPs and the number of model parameters are presented in Table~\ref{tbl:saper_FLOPs} in the main paper.

\begin{table*}[!ht]
    \centering
    \footnotesize
    \begin{tblr}{
      colspec = {l c *{8}{c}},
      column{1} = {leftsep=0pt},
      column{Z} = {rightsep=0pt},
      vline{2,3,5,7,9} = {2-Z}{dashed},   
      hline{2,4,Z} = {solid},
      hline{5,7,9,11,13} = {dashed},      
      rowsep = 1.5pt,
      colsep = 3.5pt,
      column{2} = {leftsep=6pt, rightsep=6pt},
    }
    & & \SetCell[c=4]{c}\textbf{ImageNet-1K} & & & & \SetCell[c=4]{c}\textbf{CIFAR-100} & & & \\
    & & \SetCell[c=2]{c}\textbf{ViT-S} & & \SetCell[c=2]{c}\textbf{ViT-B} & & \SetCell[c=2]{c}\textbf{ViT-S} & & \SetCell[c=2]{c}\textbf{ViT-B} & \\
    \textbf{Method} & $\boldsymbol{k}$ & \textbf{Fine-tuned} & \textbf{Distilled} & \textbf{Fine-tuned} & \textbf{Distilled} & \textbf{Fine-tuned} & \textbf{Distilled} & \textbf{Fine-tuned} & \textbf{Distilled} \\
    DINOv2 & $-$ & $80.72_{\pm 0.21}$ & N/A & $84.14_{\pm 0.32}$ & N/A & $90.81_{\pm 0.10}$ & N/A & $92.51_{\pm 0.49}$ & N/A \\
    RAPTOR & 6 & \xmark & \xmark & \xmark & \xmark & \xmark & \xmark & \xmark & \xmark \\
    \our{} & 6 & $60.33_{\pm 0.33}$ & $67.69_{\pm 1.74}$ & $58.77_{\pm 2.27}$ & $75.42_{\pm 0.54}$ & $80.41_{\pm 0.85}$ & $77.12_{\pm 0.19}$ & $78.25_{\pm 0.55}$ & $78.72_{\pm 0.13}$ \\
    RAPTOR & 12 & \xmark & \xmark & \xmark & \xmark & \xmark & $83.48_{\pm 0.52}$ & \xmark & \xmark \\
    \our{} & 12 & $68.18_{\pm 1.33}$ & $74.68_{\pm 0.11}$ & $64.33_{\pm 4.12}$ & $75.48_{\pm 2.74}$ & $\mathbf{84.48_{\pm 0.17}}$ & $83.78_{\pm 0.58}$ & $83.44_{\pm 0.44}$ & $81.52_{\pm 0.66}$ \\
    RAPTOR & 24 & \xmark & $-$ & \xmark & $\mathbf{81.2_{\pm 0.2}}$ & \xmark & $85.74_{\pm 0.21}$ & \xmark & $89.02_{\pm 0.22}$ \\
    \our{} & 24 & $74.82_{\pm 0.10}$ & $77.06_{\pm 0.14}$ & $74.83_{\pm 0.14}$ & $80.62_{\pm 0.57}$ & $\mathbf{87.57_{\pm 0.10}}$ & $86.21_{\pm 0.05}$ & $87.71_{\pm 0.34}$ & $\mathbf{89.66_{\pm 0.69}}$ \\
    RAPTOR & 36 & \xmark & $-$ & \xmark & $\mathbf{83.0_{\pm 0.1}}$ & \xmark & $86.70_{\pm 0.32}$ & \xmark & $\mathbf{90.46_{\pm 0.20}}$ \\
    \our{} & 36 & $77.47_{\pm 0.06}$ & $78.55_{\pm 0.11}$ & $77.68_{\pm 0.06}$ & $81.67_{\pm 0.15}$ & $\mathbf{88.90_{\pm 0.23}}$ & $86.91_{\pm 0.19}$ & $89.19_{\pm 0.33}$ & $90.29_{\pm 0.08}$ \\
    RAPTOR & 48 & \xmark & $-$ & \xmark & $\mathbf{83.2_{\pm 0.1}}$ & \xmark & $87.34_{\pm 0.40}$ & \xmark & $90.85_{\pm 0.10}$ \\
    \our{} & 48 & $79.17_{\pm 0.09}$ & $79.29_{\pm 0.15}$ & $79.67_{\pm 0.11}$ & $82.12_{\pm 0.31}$ & $\mathbf{89.79_{\pm 0.18}}$ & $87.56_{\pm 0.21}$ & $90.18_{\pm 0.20}$ & $\mathbf{91.39_{\pm 0.13}}$ \\
    \end{tblr}
    \caption{The overall performance of \our{} in two modes: model fine-tuning and knowledge distillation, two ViT variants: ViT-Small (ViT-S) and ViT-Base (ViT-B), and two datasets: ImageNet-1K and CIFAR-100, compared with RAPTOR. On CIFAR-100, we reproduce the authors' results, and on ImageNet-1K we report accuracy from Table 1 in the RAPTOR manuscript. Results of \our{} are averaged over 3 runs. $k$ denotes the number of heads, and the highest-performing method in each category is bolded. Results denoted by ''{\color{black}\xmark}'' are not possible to obtain since the total number of heads used by RAPTOR must be a multiple of the number of heads in a single layer, and the authors do not consider single blocks. Also, RAPTOR was not trained in a fine-tune mode. Best results in a given setting are bolded when both RAPTOR and \our{} scores are calculated. $\mathrm{N/A}$ corresponds to non-applicable cases.}
    \label{tbl:saper_extended}
\end{table*}

\subsection{Head number ablation study}\label{app:ablation_k}
\begin{figure}[!ht]
    \centering
    \includegraphics[scale=0.55]{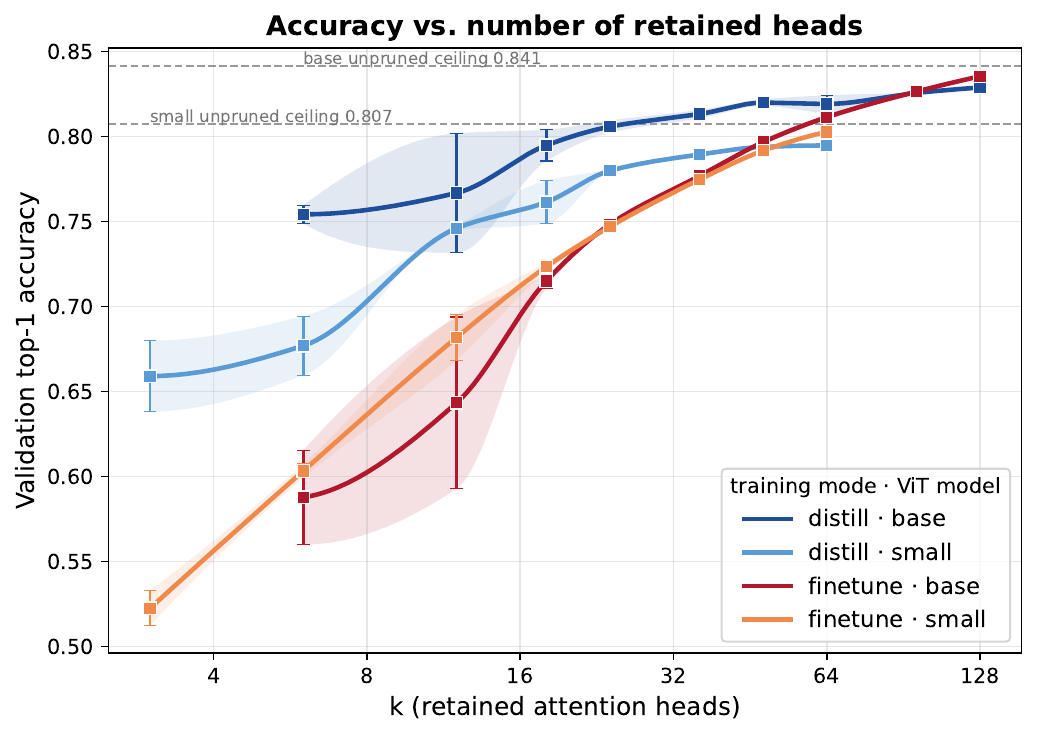}
    \caption{Accuracy against the number of retained heads $k$ on ImageNet-1K for various training modes and model sizes.\label{fig:k_sweep}} 
\end{figure}

Figure~\ref{fig:k_sweep} shows the dependence between the number of retained heads and classification accuracy for ImageNet-1K. We can observe that the fine-tuning mode is less resilient to head pruning than knowledge distillation. For ViT-S, the accuracy drops to $\approx 65\%$ in knowledge distillation, while in fine-tuning, the score is much lower ($\approx 52\%$). All considered models exceed $70\%$ accuracy for slightly more than 16 heads, and this number of heads guarantees at least $75\%$ for distilled models.

\section{Experiments setup}\label{app:experiments_setup}

We implemented \our{} in Python 3.11.5 and used following libraries: torch 2.9.1, torchvision 0.24.1, transformers 4.57.3, datasets 4.4.1, networkx 3.6.1, scikit-learn 1.8.0, pillow 12.0.0, pyarrow 22.0.0, zarr 3.1.6, numcodecs 0.16.5, numpy 2.3.5, pandas 2.3.3, huggingface-hub 0.36.0, tokenizers 0.22.1, safetensors 0.7.0, python-dotenv 1.2.1, pyyaml 6.0.3, tqdm 4.67.1, requests 2.32.5, matplotlib 3.10.8, seaborn 0.13.2, ruff 0.15.0, pytest 9.0.2, psutil 7.2.2, nvidia-ml-py 13.590.48, optuna 4.7.0, optuna-integration 4.7.0, tensorboard 2.20.0, wandb 0.25.1.

For runs on the CIFAR-100 dataset, we used NVIDIA A100 SXM4 40GB and NVIDIA H200 graphics cards available on a cluster running Slurm 26.05.2. For runs on the ImageNet-1K dataset, we used NVIDIA H200 and NVIDIA GH200 120GB graphics cards. We employed an iterative hyperparameter search strategy in which we gradually narrowed down the tuning ranges of the hyperparameters or set some of them to fixed values, which were found to be best in previous tuning iterations.

\subsection{Hyperparameters}

Some of the hyperparameters that we used included:
\begin{description}
  \item[batch\_size]: the mini-batch size - if not specified otherwise, we used a value of 256.
  \item[alpha\_init]: the value of the $\alpha$ hyperparameter of LapSum, annealed from \emph{alpha\_init} to \emph{alpha\_final} (which was fixed at 0.01) over \emph{alpha\_anneal\_epochs} using \linebreak \emph{alpha\_anneal\_schedule} schedule.
  \item[alpha\_anneal\_epochs]: the number of epochs over which LapSum $\alpha$-annealing is performed.
  \item[alpha\_anneal\_schedule]: the schedule used for $\alpha$-annealing, one of: 'cosine', 'geometric', 'linear'.
  \item[annealing\_start\_coeff]: controls when the LapSum $\alpha$-annealing starts, calculated as
  \begin{equation*}
      round\left(\mathrm{annealing\_start\_coeff} \cdot \mathrm{(epochs - alpha\_anneal\_epochs)}\right).
  \end{equation*}
  \noindent This hyperparameter may be tuned independently from \emph{alpha\_anneal\_epochs}.
  \item[k]: the number of heads to be retained after pruning.
  \item[k\_warmup\_epochs]: the number of initial epochs over which we linearly reduce the $k$ hyperparameter, from all the heads (72 for ViT-S, 144 for ViT-B) to the value specified by $k$ (set to 24 for all the Optuna tuning runs).
  \item[score\_init]: initialization method for LapSum head importance scores; (1) $0.01 \cdot \mathcal{N}(0, 1)$, (2) $\mathcal{N}(0, 1)$ and (3) $\frac{\mathcal{N}(0, 1)}{\sqrt{n}}$. The default value for most runs was $0.01 \cdot \mathcal{N}(0, 1)$.
  \item[weight\_decay]: L2 regularization hyperparameter, applied to all trained model parameters apart from LapSum selector scores. If not specified otherwise, we use $2 \cdot 10^{-5}$.
  \item[selector\_grad\_clip\_norm]: used for gradient norm clipping only for the LapSum selector score gradients. In all experiments, we set it to $5$.
  \item[backbone\_freezing\_strategy]: specifies which parameters of the backbone network are trainable. '\underline{unfreeze\_all}' means that all the backbone parameters are trainable, \linebreak '\underline{freeze\_attention}' means all attention parameters (query, key and values) are frozen while the rest of backbone is trainable, '\underline{freeze\_attention\_and\_mlp}' is similar but, apart from attention parameters, also MLP weights at the end of each layer are left as untrainable and '\underline{freeze\_all}' means that the whole backbone network is untrainable. In all cases, the linear classifier is trainable, and for the distillation runs the student network projections are always trainable.
  \item[backbone\_lr]: the initial learning rate used for training the backbone parameters. Along with other learning rate hyperparameters, it is annealed using \emph{lr\_scheduler} (for all tuning runs it was held fixed at 'cosine') from the initial learning rate to a value calculated as
  \begin{equation*}
      \mathrm{lr\_scheduler\_param} \cdot min \lbrace \mathrm{backbone\_lr}, \mathrm{selector\_lr}, \mathrm{lr} \rbrace.
  \end{equation*} 
  \item[selector\_lr]: the initial learning rate for the LapSum head selector parameters.
  \item[lr]: the default learning rate, used for all parameters apart from backbone and selector parameters. In practice, this means that it is used for the linear classifier in the fine-tune mode, and for the student network projections as well as for the final linear probe (not used in the tuning runs) in the distillation mode.
  \item[lr\_scheduler\_param]: used as a hyperparameter for the \emph{lr\_scheduler} and has different meaning depending on the selected scheduler. For 'exponential' and 'step' learning rate schedulers, \emph{lr\_scheduler\_param} is used for the $\gamma$ parameter, with common values such as 0.95 or 0.99. For 'cosine', \emph{lr\_scheduler} is used to calculate the final learning rate \emph{eta\_min}.
  \item[distill\_layers]: a set of layers for which we compute the distillation loss against the same layer of the teacher network; example value: $\lbrace 4,8,-1 \rbrace$, where $-1$ identifies the last feature extractor layer and numbering of layers starts from one.
  \item[distill\_loss]: the distillation loss, one of 'cosine', 'kl' (Kullback-Leibler divergence) or 'mse' (mean squared error).
  \item[kl\_temp]: used only with Kullback-Leibler divergence loss; it sets the temperature used for softening the student and teacher activations by dividing them before the softmax application.
  \item[distill\_token\_pool]: defines which tokens are used for the distillation: 'all' - the whole hidden state of a layer, 'cls' - only the CLS token, 'cls\_mean' - the CLS token and the mean of the patch tokens, kept as two tokens.
  
  \item[teacher\_forcing\_mode]: when active, the backbone network is effectively divided into subsegments defined by \emph{distill\_layers}. In such a case, at each specified layer, we calculate the distillation loss by comparing the teacher activations (denoted as '$teacher$') with the corresponding activations of the student network ('$student$'). Then, we overwrite the student activations for that layer, according to one of the modes. The 'mix' mode combines teacher and student activations using
  \begin{equation*}
    \lambda \cdot teacher + (1-\lambda) \cdot student,
  \end{equation*}
  while 'sample' mode means that with probability $\lambda$, we select only the teacher activations per given sample, and only the student activations otherwise. The $\lambda$ parameter is cosine annealed through \emph{teacher\_forcing\_anneal\_epochs} from \emph{teacher\_forcing\_alpha\_start} to 0.
\end{description}

\subsection{Hyperparameter tuning on CIFAR-100}

We tuned the hyperparameters for ViT-S and ViT-B variants separately on CIFAR-100, in two training modes: fine-tuning with a classification head and knowledge distillation, in which we used kNN for evaluation and trial scoring. Each trial consisted of 25-40 epochs of soft-mask training followed by 10-40 epochs of hard-mask training in which the heads selected for pruning were removed from the network.

For fine-tuning of ViT-S, we executed a total of 235 trials in three hyperparameter optimization iterations. We then proceeded to hyperparameter tuning for knowledge distillation of ViT-S, in which we executed 269 trials, with the last 75 trials focused on checking whether performing distillation on the output of the Multi-Head Attention block is better than distillation at the output of the layer (after MLP), as well as on checking whether teacher forcing is beneficial. In these runs, the \emph{teacher\_forcing\_anneal\_epochs} was fixed to 5, and the \emph{teacher\_forcing\_alpha\_start} was set to 0.5. We found that distillation at the MLP layer output and without teacher forcing provide better results.
Following these runs were 159 trials for fine-tuning of ViT-B, in which we set the batch size to 64, which had an impact on learning rate values, set the \emph{alpha\_anneal\_schedule} hyperparameter to 'geometric' and score\_init to $\frac{\mathcal{N}(0, 1)}{\sqrt{n}}$. The last run consisted of 62 trials for the knowledge distillation of ViT-B, in which we returned to a batch size of 256. The summary of the explored hyperparameter space and the best hyperparameters is presented in Table~\ref{tab:hyper_cifar}.

\begin{table*}[!ht]
    \centering
    \footnotesize
    \caption{The set of considered and best hyperparameters on CIFAR-100. Values marked in \textcolor{blue}{blue} were not optimized and instead were set as fixed for a given configuration.}
    \label{tab:hyper_cifar}
    \vspace{0.5em}
    \begin{tblr}{
      colspec = {l Q[l,wd=4.5cm] c c c c},
      hline{1,Z}  = {solid},         
      hline{2} = {3}{solid, leftpos=-1, rightpos=1},   
      hline{2} = {4}{solid, leftpos=1, rightpos=-1},   
      hline{2} = {5}{solid, leftpos=-1, rightpos=1},     
      hline{2} = {6}{solid, leftpos=1, rightpos=-1},     
      hline{3}    = {solid},         
      hline{8,13} = {dashed},        
      rowsep = 1.5pt,
      colsep = 4pt,
    }
    & & \SetCell[c=2]{c} \textbf{Fine-tuning} & & \SetCell[c=2]{c} \textbf{Distillation} \\
    \textbf{Hyperparameter} & \textbf{Search range} & \textbf{ViT-S} & \textbf{ViT-B} & \textbf{ViT-S} & \textbf{ViT-B} \\
        \texttt{alpha\_init}  & \{1, 3, 5\}   & \textcolor{blue}{1} & 1 & \textcolor{blue}{1} & \textcolor{blue}{1} \\
        \texttt{alpha\_anneal\_epochs}        & [0, 40] & 0 & 2 & 2 & 4 \\
        \texttt{alpha\_anneal\_schedule}      & \{cosine, geometric, linear\} & cosine & \textcolor{blue}{cosine} & \textcolor{blue}{geometric} & \textcolor{blue}{cosine} \\
        \texttt{annealing\_start\_coeff}      & [0, 1] & 0.327 & 0.518 & 0.091 & 0.618 \\
        \texttt{k\_warmup\_epochs}            & [0, 20] & 10 & 5 & 9 & 0 \\
        \texttt{backbone\_freezing\_strategy} & \{unfreeze\_all, freeze\_attention, freeze\_attention\_and\_mlp, freeze\_all\} & unfreeze\_all & freeze\_attention & \textcolor{blue}{freeze\_attention}   & \textcolor{blue}{freeze\_attention} \\
        \texttt{backbone\_lr}                 & $[10^{-7}, 10^{-3}]$ & $5 \cdot 10^{-5}$ & $1.079 \cdot 10^{-4}$ & $2.654 \cdot 10^{-5}$ & $3.019 \cdot 10^{-5}$ \\
        \texttt{selector\_lr}                 & $[10^{-5}, 10^{-2}]$ & $9.9 \cdot 10^{-3}$ & $7.493 \cdot 10^{-3}$ & $1.598 \cdot 10^{-3}$ & $5.068 \cdot 10^{-4}$ \\
        \texttt{lr}                           & $[10^{-6}, 10^{-2}]$ & $7.766 \cdot 10^{-5}$ & $\textcolor{blue}{10^{-3}}$ & $2.807 \cdot 10^{-3}$ & $\textcolor{blue}{10^{-3}}$ \\
        \texttt{lr\_scheduler\_param}         & \{0.1, 0.033333, 0.01, 0.0\} & 0.1 & \textcolor{blue}{0.1} & \textcolor{blue}{0.1} & \textcolor{blue}{0.1} \\
        \texttt{distill\_layers}              & \{$\text{all\_layers}$, $\text{'4,8,last'}$, $\text{'8,last'}$, $\text{'last'}$\} & - & 'last' & - & $\text{'8,last'}$ \\
        \texttt{distill\_loss}                & \{cosine, kl, mse\} & - & cosine & - & \textcolor{blue}{cosine} \\
        \texttt{kl\_temp}                     & $[1, 4]$ & - & 1 & - & - \\
        \texttt{distill\_token\_pool}         & \{all, cls, cls\_mean\} & - & cls\_mean & - & \textcolor{blue}{cls\_mean} \\
        \texttt{teacher\_forcing\_mode}       & \{mix, sample, none\} & - & mix & - & - \\
    \end{tblr}
\end{table*}

\subsection{Hyperparameter tuning on ImageNet-1K}

For hyperparameter tuning on ImageNet-1K, we utilized caches of processed images and teacher activations to speed up each run. In each epoch, we processed a random subset of 64,000 examples, not the entire training set, in contrast to CIFAR-100 experiments. The soft-mask training phase consisted of 20-30 epochs, followed by 10 epochs of hard-mask training. In all runs, the \emph{backbone\_freezing\_strategy} was fixed to 'freeze\_attention', \emph{distill\_loss} to 'cosine', \emph{distill\_token\_pool} to 'cls\_mean' and \emph{lr\_scheduler\_param} to 0.1.
We executed 205 trials for fine-tuning of ViT-S, 206 trials for fine-tuning of ViT-B and a total of 639 trials for distillation of ViT-S and ViT-B.
The summary of the considered hyperparameters, their ranges and optimal sets is presented in Table~\ref{tab:hyper_imagenet}.

\begin{table*}[!ht]
    \centering
    \small
    \caption{The set of considered and best hyperparameters on ImageNet-1K.}
    \label{tab:hyper_imagenet}
    \vspace{0.5em}
    \begin{tblr}{
      colspec = {l Q[l,wd=4.5cm] c c c},
      hline{1,Z}  = {solid},         
      hline{2} = {3}{solid, leftpos=-1, rightpos=1},   
      hline{2} = {4}{solid, leftpos=1, rightpos=-1},   
      hline{2} = {5}{solid, leftpos=-1, rightpos=-1},     
      hline{3}    = {solid},         
      hline{8,13} = {dashed},        
      rowsep = 1.5pt,
      colsep = 4pt,
    }
    & & \SetCell[c=2]{c} \textbf{Fine-tuning} & & \SetCell[c=1]{c} \textbf{Distillation} \\
    \textbf{Hyperparameter} & \textbf{Search range} & \textbf{ViT-S} & \textbf{ViT-B} & \textbf{ViT-S \& ViT-B} \\
    \texttt{alpha\_init}            & {[0.5, 100.0]}                            & 3.56261               & 1.26986               & 18.2912 \\
    \texttt{alpha\_final}           & {$[10^{-4}, 10^{-1}]$}                    & $2.283 \cdot 10^{-3}$ & $4.449 \cdot 10^{-4}$ & $9.904 \cdot 10^{-3}$ \\
    \texttt{alpha\_anneal\_epochs}  & {[0, 30]}                                 & 9                     & 1                     & 18 \\
    \texttt{alpha\_anneal\_schedule} & {\{cosine, geometric, linear\}}          & geometric             & geometric             & geometric \\
    \texttt{annealing\_start\_coeff} & {[0, 1]}                                 & 0.0284836             & 0.487653              & 0.247263 \\
    \texttt{weight\_decay}          & {$[10^{-6}, 10^{-3}]$}                    & $2.246 \cdot 10^{-6}$ & $1.565 \cdot 10^{-6}$ & $2.072 \cdot 10^{-6}$ \\
    \texttt{k\_warmup\_epochs}      & {[0, 16]}                                 & 4                     & 8                     & 6 \\
    \texttt{backbone\_lr}           & {$[5 \cdot 10^{-7}, 5 \cdot 10^{-4}]$}    & $2.884 \cdot 10^{-5}$ & $3.630 \cdot 10^{-5}$ & $7.625 \cdot 10^{-5}$ \\
    \texttt{selector\_lr}           & {$[10^{-3}, 0.5]$}                          & $4.489 \cdot 10^{-2}$ & $5.936 \cdot 10^{-3}$ & $9.189 \cdot 10^{-2}$ \\
    \texttt{lr}                     & {$[5 \cdot 10^{-4}, 0.5]$}                & $1.062 \cdot 10^{-3}$ & $9.462 \cdot 10^{-4}$ & $6.789 \cdot 10^{-3}$ \\
    \texttt{distill\_layers}        & {\{$\text{all\_layers}$, $\text{'8,last'}$, $\text{'last'}$\}} & $-$ & $-$ & $\text{'8,last'}$ \\
    \end{tblr}
\end{table*}

\end{document}